\documentclass[11pt]{article}

\usepackage[final]{acl}

\usepackage{times}
\usepackage{latexsym}

\usepackage[T1]{fontenc}
\usepackage[utf8]{inputenc}

\usepackage{microtype}

\usepackage{inconsolata}

\usepackage{graphicx}

\usepackage{tikz}
\usepackage{amsmath}

\usepackage{booktabs}
\usepackage{tabularx}
\usepackage{multirow}

\usepackage{xcolor}

\usepackage{listings}
\usepackage[most]{tcolorbox}

\usepackage[caption=false,font=footnotesize]{subfig}

\newcommand{\mypara}[1]{\noindent\textbf{#1}.}

\newcommand{\new}[1]{#1}



\lstdefinelanguage{json}{
  morestring=[b]",
  morecomment=[l]{//},
  morecomment=[s]{/*}{*/},
  morekeywords={true,false,null}
}

\lstdefinestyle{prompt}{%
  basicstyle=\ttfamily\small,
  breaklines=true,
  frame=single,
  backgroundcolor=\color{gray!5},
  columns=fullflexible,
  showstringspaces=false,
  upquote=true,
  aboveskip=0.8\baselineskip,
  belowskip=0.8\baselineskip,
  abovecaptionskip=0.4\baselineskip,
  belowcaptionskip=0.4\baselineskip,
  xleftmargin=0.5em,
  xrightmargin=0.5em,
  inputencoding=utf8,
  extendedchars=true,
  literate={–}{{--}}1{—}{{---}}1{'}{{'}}1{"}{{``}}1{"}{{''}}1{…}{{...}}1
}
\title{MemeTrace: A Multimodal LLM-Based Agent for Safety-Aware Meme Provenance Analysis}

\author{ Bo Shao,  Yihan Ma,  Yiting Qu,  Michael Backes,  Yang Zhang\thanks{Corresponding author.} \\
CISPA Helmholtz Center for Information Security \\
\texttt{\{bo.shao, yihan.ma, yiting.qu, director, zhang\}@cispa.de}
}

\begin{document}
\maketitle

\begin{abstract}
Online memes are often carriers of unsafe content such as disinformation and propaganda.
Their rapid evolution in response to news events and strong dependence on contextual information make it difficult for platforms and moderators to detect and moderate them effectively.
Moreover, existing moderation studies typically classify memes as safe or unsafe based on individual images, lacking the ability to perform broader \textbf{provenance analysis}, i.e., where a meme originates, how it spreads, what cultural symbolism it represents, whether it poses potential risks to society, etc.
To bridge this gap, we introduce \textbf{MemeTrace}, a multimodal LLM-based agent that automatically traces the provenance and potential risks of emerging memes.
We evaluate MemeTrace on Meme100, a dataset of 100 real-world memes, using both human annotation and automatic metrics.
The results demonstrate that MemeTrace substantially outperforms the VLM baseline.
More importantly, \textbf{MemeTrace} achieves higher accuracy in detecting unsafe memes, especially those involving disinformation and sensitive content, and remains robust under visual perturbations. 

\end{abstract}
\section{Introduction}
\begin{figure}[t]
  \centering
    \includegraphics[width=\columnwidth]{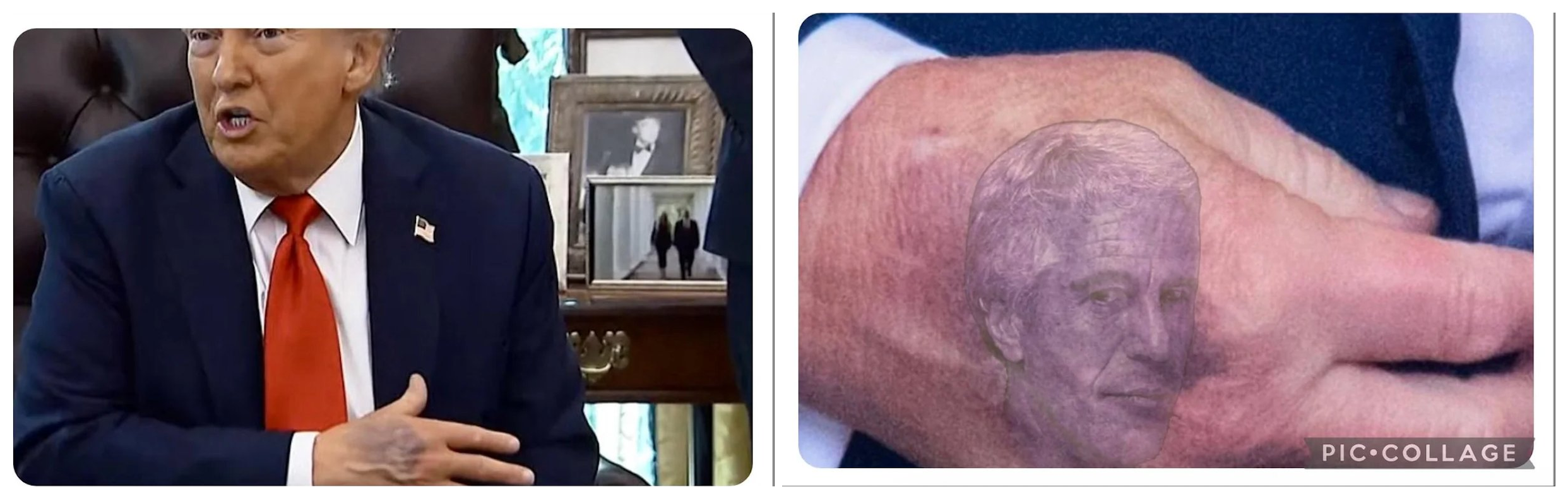}
    \caption{Meme: Donald Trump with a bruise overlaid with Jeffrey Epstein's face, invoking conspiracy theories about their alleged associations.}
  \label{fig:image88}
\end{figure}

Memes are sets of digital items, such as images, videos, and phrases, that spread rapidly across social media platforms~\cite{memes,B11,ZCBCSSS18}.
While most memes are benign, some carry unsafe content such as hateful ideology~\cite{Y16, QHPBZZ23, QSHBZZ23}, misinformation~\cite{KB24}, and conspiracy theories~\cite{S25}.
Given their broad reach and rapid dissemination, content moderation is critical.
However, interpreting the real intent of memes is inherently challenging for two reasons.
First, memes are often complex combinations of visual elements, textual phrases, hidden cultural references, and the creator's attitude~\cite{memes,B11}.
Rich background knowledge is often required to interpret memes correctly and to determine whether they might be harmful~\cite{MSQYBZZ25}.
Second, the intent and implication a meme conveys can be implicit~\cite{ZCBCSSS18, MSQYBZZ25}.
Humor, irony, and satire can often mask a meme's true intent, making it difficult to identify implicit harmful content.
For example, \autoref{fig:image88} shows Donald Trump with a bruise overlaid with Jeffrey Epstein's face, invoking conspiracy theories about their alleged associations.

Previous research on memes mainly focuses on closed-set classification tasks (e.g., hateful vs. benign)~\cite{LHCRASY20,ZC202,PSDANC21,SBSPDCPG20}. 
These methods mainly rely only on surface-level visual or textual elements, without considering contextual information.
As a result, they provide limited insights into why a meme is hateful or harmful.
More importantly, these methods often exhibit poor generalizability to newly emerging memes due to lack of up-to-date background knowledge, e.g., recent news, events, or platform-specific usage.
For such context-dependent data, \emph{provenance analysis}, namely the examination of the data origin, reuse history, and other contextual information, is required to reveal whether and why the data may be hateful or harmful.
This motivates us to develop a meme analysis framework that supports provenance analysis at scale by leveraging dynamic external knowledge and step-by-step reasoning, moving beyond static, closed-set classification.

\mypara{Our Work}
In this paper, we introduce and formalize a new task, \textbf{Meme Provenance}, a type of provenance analysis tailored specifically to real-world memes.
Given a meme image, the goal is to generate a structured analytical report across 11 fields namely about, origin, spread, culture, structure, usage, target audience, emotional impact, cognitive load, persuasive technique, and safety flags.
To achieve this goal, we design \textbf{MemeTrace}, a multimodal LLM-based agent that performs evidence-driven meme provenance analysis from a single image input.
Instead of relying on one-shot VLM inference, MemeTrace follows a Think--Act--Evaluate loop that coordinates visual perception, external retrieval, evidence assessment, and report synthesis.

We evaluate MemeTrace on \textbf{Meme100}, a new dataset of 100 real-world memes collected from Reddit, Twitter/X, 4chan, Facebook, TikTok, YouTube, and Instagram, covering both classic and newly emerged memes across multiple topics.
For each meme, MemeTrace generates one provenance report, which we evaluate through three-expert human evaluation on a five-point Likert scale~\cite{L32}.
Apart from human evaluation, we further propose a suite of automatic metrics, derived directly from the generated reports to assess their overall quality.
Using these evaluations, we compare MemeTrace with the GPT-4o VLM baseline~\cite{GPT4o}, and further assess its robustness under image perturbations and backbone transferability with alternative VLMs.

\mypara{Results}
The evaluation result shows that MemeTrace substantially outperforms the VLM-only baseline, in terms of comprehensively and accurately interpreting real-world memes, including identifying the implicit hateful content.
Through careful ablation studies, we find that the specialized tools, especially Tavily and reverse image search, contribute most to MemeTrace's performance. 
This indicates that external contextual information is critical for analyzing real-world memes and determining whether their underlying intent is hateful.
MemeTrace also demonstrates strong robustness under common Internet image perturbations, such as Gaussian blur, UI overlay, and border cropping.
Furthermore, by replacing the backbone with alternatives like Qwen2.5-VL-72B and Gemini-Pro-1.5~\cite{BCLWGSDWWTZZYLWWDFXYZXCZYXL25,RSTLLASLFSAABDMDGSLVRXMCIBHMJSCRMAGMTYMASAIKLHSSCRSSo24}, we show that MemeTrace achieves strong backbone transferability, suggesting that its performance gains come from the agent design rather than a specific VLM.

Overall, we make the following contributions:

\begin{itemize}
  \item We introduce meme provenance as a new task and define a provenance report format for understanding real-world memes.
  \item We present MemeTrace, an LLM-based agent that uses local database retrieval, reverse image search, and targeted web search to synthesize structured provenance reports.
  \item We curate Meme100, a dataset of 100 real-world memes. MemeTrace significantly outperforms the VLM baseline and demonstrates strong robustness under image perturbations.
  \item We develop a suite of automatic report-quality metrics that show high correlation with human scores, enabling scalable evaluation.
\end{itemize}

\section{Related Work \& Problem Statement}
\label{section: related_work_problem_statement}

\begin{figure*}[t]
  \centering
  \includegraphics[width=\textwidth]{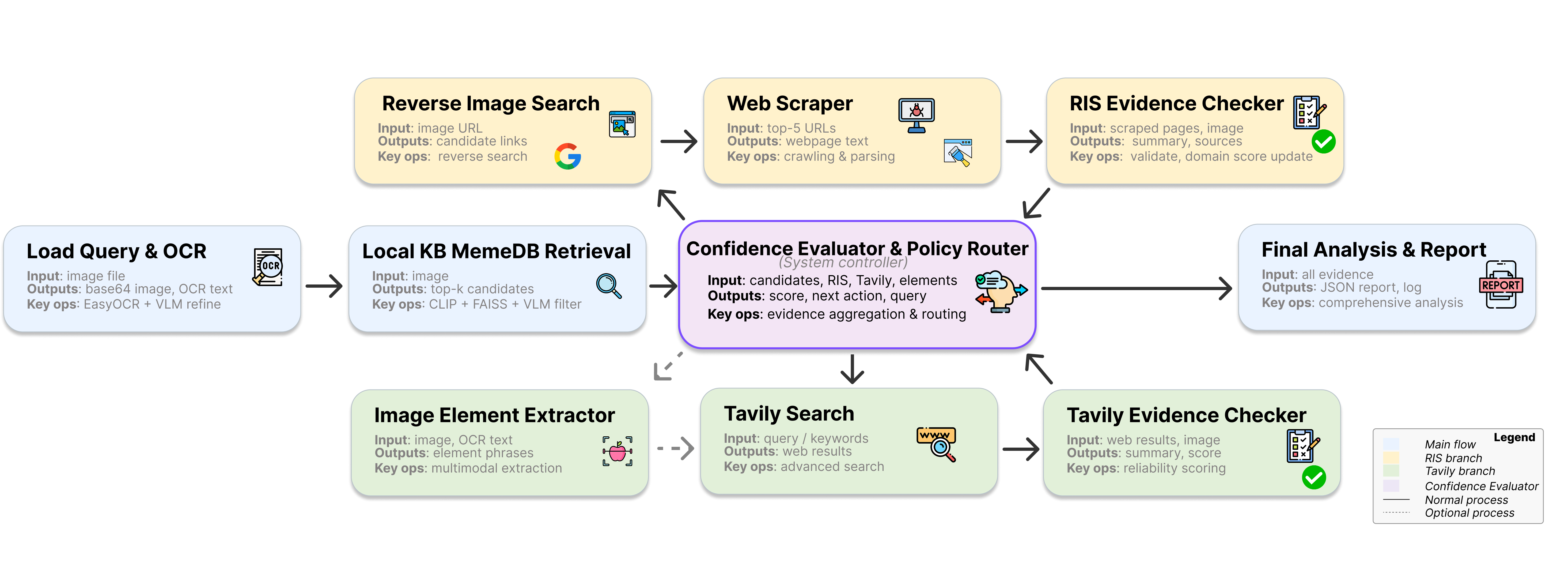}
  \caption{Overview of the MemeTrace agent architecture, illustrating the multimodal perception, evidence retrieval, and confidence-routed provenance reasoning loop.}
  \label{fig:overview}
\end{figure*}

\mypara{Related Work}
Existing studies on meme analysis generally focus on narrow, single-dimensional analytical tasks. 
One important line of meme studies~\cite{M20, KFMGSRT20, SBSPDCPG20, KFMGSFBLNZMVRLHCRASYSOPSP20, PDMSANC21, PSDANC21, CLCJ22, CHKCLJ23} utilizes multimodal models to classify memes as hateful, offensive, or harmful. 
For example, \emph{SemEval Memotion}~\cite{SBSPDCPG20} uses fused features from a CNN for images and a BiLSTM for text to predict each meme’s sentiment, e.g., whether the meme contains offensive content. 
Similarly, \emph{MOMENTA}~\cite{PSDANC21} fuses image features and overlaid text features for harmful meme detection. 

Another line of meme studies~\cite{ZCBCSSS18, LABCZS21, QHPBZZ23, QSHBZZ23, BHHH23, MSQYBZZ25, LLWYM24} shifts the focus toward a deeper understanding beyond the meme’s surface elements, such as meme genealogy~\cite{ZCBCSSS18, LABCZS21}, evolution analysis~\cite{QHPBZZ23}, and the sociopolitical ideologies that memes represent~\cite{BHHH23, MSQYBZZ25, LLWYM24}. 
But these works still rely on static multimodal models and focus on only one analytical dimension at a time, lacking the ability to incorporate real-world evidence or broader online contextual knowledge. 

To address these limitations, recent works~\cite{LLGMWY24, LFLWD25, HLLLCM24, HPWLL25} have started exploring LLM-based agents for meme analysis.
However, these methods remain limited to single-meme classification tasks and do not provide systematic modeling of the broader background of memes.

\mypara{Problem Statement}
In this study, we aim to decompose a complex meme into a detailed and interpretable analytical report.
The objective is to design a system that, given a target meme image, automatically generates a structured analytical report capturing its semantic content, spreading dynamics, cultural context, psychological influence, and safety assessment.
To make these aspects concrete and measurable, we further decompose them into 11 analytical fields: \textit{About}, \textit{Structure}, \textit{Origin}, \textit{Spread}, \textit{Culture}, \textit{Usage}, \textit{Target Audience}, \textit{Emotional Impact}, \textit{Cognitive Load}, \textit{Persuasive Technique}, and \textit{Safety Flag} (as summarized in Appendix~\ref{appendix:provenance_fields}).

\section{The MemeTrace Agent}

\subsection{Design Rationale}

To conduct meme provenance analysis, a natural solution is to leverage vision-language models (VLMs) such as GPT-4V~\cite{GPT-4V} to interpret the semantics of a new input meme. 
However, these models are usually static and cannot incorporate up-to-date knowledge, which limits their adaptability to the rapidly evolving online environment.
As a result, they can only recognize outdated surface-level elements, such as visual cues or overlaid text, while failing to capture deeper and newer meaning, cultural context, and linked background knowledge.
To overcome this limitation, we aim to design a multimodal LLM-based agent that automatically collects up-to-date online information about a meme, integrates and interprets content from diverse sources, and verifies the reliability and authenticity of the gathered information.
With the verified information about the meme (e.g., semantics, spread, and origin), the agent automatically synthesizes a comprehensive analytical report for meme provenance.

\subsection{System Overview}
\label{sec:architecture}

MemeTrace is a multimodal LLM-based agent that performs evidence-driven meme provenance analysis from a single image input. 
Instead of generating a report from a one-shot VLM inference, MemeTrace follows a Think–Act–Evaluate loop that integrates perception, retrieval, and reasoning through coordinated modules.
\autoref{fig:overview} shows the overview of MemeTrace, which comprises six core modules operating in a closed loop to enable end-to-end meme provenance analysis.

Given an input meme, MemeTrace first extracts the overlaid texts through the OCR module, then checks whether the meme (or a visually similar one) already exists in the local meme database via the Local Database Retrieval module.
The retrieved evidence is then evaluated by the confidence evaluator, which assesses whether the evidence across the eleven fields required for meme provenance analysis is sufficient and trustworthy.
If the evidence is incomplete or uncertain, the confidence evaluator activates the Reverse Image Search (RIS) and Tavily Search modules sequentially.
It generates context-aware query keywords to collect up-to-date information about the meme from the web, based on both its visual features (meme image) and extracted OCR texts.
The newly gathered RIS and Tavily results are re-evaluated by the confidence evaluator to determine evidence sufficiency and consistency.
If inconsistencies remain, the confidence evaluator refines its queries and iteratively retrieves new evidence through Tavily Search until all required evidence is considered sufficient and trustworthy for generating the final meme provenance report.
Implementation details are provided in Appendix~\ref{appendix:implementation}.

\begin{figure}[t]
\centering
\begin{lstlisting}[basicstyle=\ttfamily\tiny, breaklines=true, showstringspaces=false]
{
  "new_confidence_score": 0.85,
  "next_action": "GENERATED_QUERY_SEARCH",
  "justification": "The evidence is largely consistent, with detailed accounts of the real-life incident involving Alison Whelan...",
  "next_query": "Alison Whelan ferry meme Reddit spread"
}
\end{lstlisting}
\caption{Example output of the confidence evaluator.}
\label{fig:confidence-eval-example}
\end{figure}

\subsection{Core Modules}
\label{sec:core-modules}

\subsubsection{Image Loading \& OCR Extraction}
This module loads the input meme image and extracts English text from the image using the EasyOCR library~\cite{easyocr}. 
Although OCR functions are mature, the raw output often contains irrelevant content due to complex meme images, such as watermarks. 
To remove them, the system uses a multimodal LLM (GPT-4o)~\cite{GPT4o} to double-check the results, correcting errors and preserving the meme’s core textual content. 
The resulting text provides textual clues for downstream modules in the provenance analysis pipeline. 

\subsubsection{Local Database Retrieval}
This module identifies similar meme templates and retrieves their background information from a local database.
We maintain \textbf{MemeDB}, containing 21K+ confirmed memes from KnowYourMeme~\cite{kym}.
Each meme's image and metadata are encoded using CLIP~\cite{RKHRGASAMCKS21}, concatenated, and indexed with FAISS~\cite{DGDJSMLHJ26} for fast similarity search.
Given an input meme, the system queries the index to retrieve the top-3 similar candidates.
Details in Appendix~\ref{appendix:memedb}.

\subsubsection{Reverse Image Search (RIS) Branch}
The RIS branch retrieves webpages related to the input meme using Google RIS~\cite{google_images}.
To prioritize credible sources, we rank retrieved webpages using a domain-based scoring algorithm (Appendix~\ref{appendix:domain_scoring}) and crawl only the top-5.
The crawled content is then processed by the RIS evidence checker, which uses GPT-4o to extract structured summaries with inline citations and assigns a usefulness score (0--1) to each source.
These scores are used to update domain rankings dynamically, enabling the system to adapt to source quality over time.
Details in Appendix~\ref{appendix:ris-example}.

\subsubsection{Open-Web Search (Tavily) Branch}
The Tavily branch uses the external search engine Tavily to collect missing background knowledge related to the meme. 
It operates in two complementary modes based on the current evidence state, i.e., whether the collected evidence is sufficient and trustworthy.
In \textbf{Element-Extraction Mode}, the agent performs broad exploratory search when reverse image search provides insufficient evidence. 
It identifies salient visual and textual elements in the meme, especially proper nouns such as people, organizations, or places, and combines the top two ranked elements into a Tavily query. 
This retrieves up to five webpage results with short snippets and source links, helping the agent discover potentially relevant context.
In \textbf{Evaluator-Query Mode}, the agent performs targeted evidence completion when partial evidence is available but some fields remain uncertain, incomplete, or inconsistent. 
The confidence evaluator formulates a query based on the current evidence state, and the Tavily quality checker summarizes the retrieved results with citations and quality scores (0--1), which guide the evaluator's next actions.

The two complementary modes support a feedback-driven reasoning loop where quality scores and summaries guide the evaluator to refine subsequent queries and avoid low-quality search directions.

\subsubsection{Confidence Evaluator \& Routing Policy}
\label{sec:evaluator}
The confidence evaluator acts as the ``brain'' of MemeTrace, powered by a multimodal LLM with a carefully designed prompt.
After each tool invocation, it assesses all collected evidence to determine whether the information is sufficient or conflicting, and decides whether to invoke additional retrieval modules, generate new queries, or proceed to report synthesis.
As illustrated in \autoref{fig:overview}, the evaluator is first activated after local database retrieval.
If the retrieved information is insufficient, it triggers the RIS branch.
Once RIS results are obtained, the evaluator reassesses evidence sufficiency.
If evidence remains incomplete, it proceeds to the Element-Extraction Mode within the Tavily branch to dig for more potential information.
When evidence is still insufficient, the evaluator activates the Evaluator-Query Mode, where it refines search keywords based on all prior evidence (local database, RIS, and Tavily history) and retrieves iteratively until evidence is sufficient or the maximum iteration limit is reached.
\autoref{fig:confidence-eval-example} shows an example output of the confidence evaluator. 

\subsubsection{Report Synthesis}
\label{sec:report}

Once the confidence evaluator determines that evidence is sufficient or the iteration limit is reached, the report synthesis module uses GPT-4o with a carefully designed prompt to integrate all collected evidence into a structured meme provenance report following the schema in Section~\ref{section: related_work_problem_statement}.
An example output is shown in \autoref{fig:case_img33_immigration}.

\section{Experimental Setup}

\subsection{Evaluation Datasets}
\label{sec:datasets}

We employ three datasets: the main dataset \textbf{Meme100}, and two derived subsets (\textbf{ablation subset} and \textbf{robustness subsets}) for ablation and robustness testing.

\mypara{Meme100}
We construct \textbf{Meme100}, a dataset of 100 memes including both classic templates (from KnowYourMeme) and newly emerged memes collected from social platforms (Reddit, Twitter/X, 4chan, Facebook, TikTok, YouTube, Instagram) using an event-driven strategy.
Newly emerged memes are associated with high-impact events (e.g., 2024 U.S. election, Russia–Ukraine conflict).
The dataset spans diverse topics (politics, technology, social trends), modalities (image-only, text-heavy), and platforms.
Full collection methodology and distribution statistics  are provided in Appendix~\ref{appendix:meme100}.

\mypara{Ablation Subset}
We extract 20 samples from Meme100, maintaining distributions across topic, platform, sparsity, and modality consistent with the main set.
This subset is used to quantify the contribution of individual modules and to evaluate the impact of replacing the multimodal LLM backbone (Qwen2.5-VL-72B, Gemini-Pro-1.5) under identical settings.

\begin{figure}[t]
  \centering

  \begin{minipage}[t]{0.38\linewidth}
    \centering
    \includegraphics[height=2.6cm]{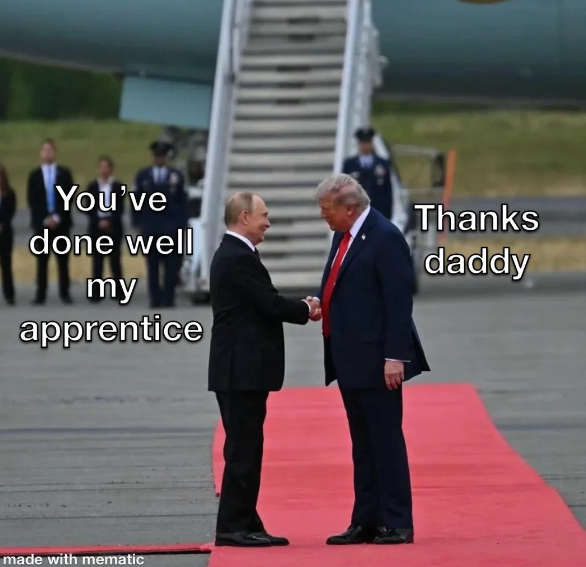}
    {\scriptsize \mbox{(a) Original meme}}
  \end{minipage}
  \hfill
  \begin{minipage}[t]{0.38\linewidth}
    \centering
    \includegraphics[height=2.6cm]{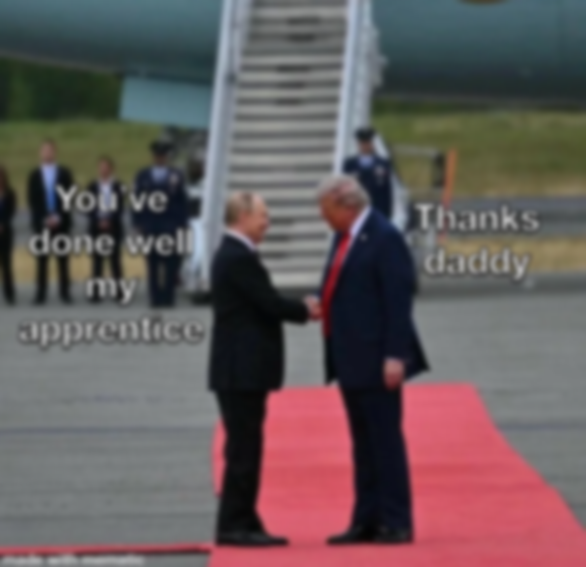}
    {\scriptsize \mbox{(b) Gaussian blur ($\sigma$=2.5)}}
  \end{minipage}

  \begin{minipage}[t]{0.38\linewidth}
    \centering
    \includegraphics[height=2.6cm]{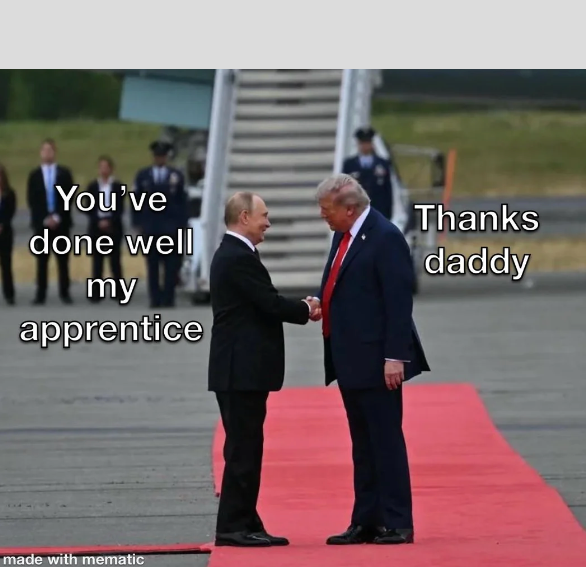}
    {\scriptsize \mbox{(c) UI overlay (12\%)}}
  \end{minipage}
  \hfill
  \begin{minipage}[t]{0.38\linewidth}
    \centering
    \includegraphics[height=2.6cm]{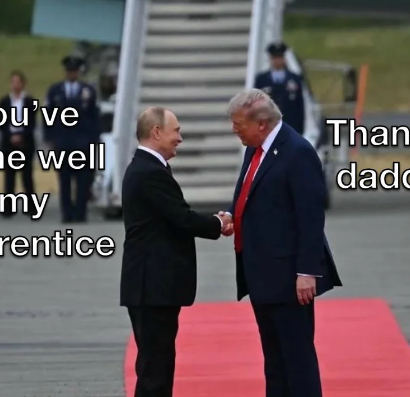}
    {\scriptsize \mbox{(d) Border cropping (15\%)}}
  \end{minipage}

  \caption{
  Examples of three perturbation types. The figure shows the original meme (top left) and its three perturbed variants. 
  } 
  \label{fig:perturbation_examples}
\end{figure}

\mypara{Robustness Subsets}
To evaluate system robustness under realistic visual perturbations, we construct three robustness subsets based on the ablation subset through image transformations: Gaussian blur ($\sigma = 2.5$), border cropping (15\%), and UI overlays ($\approx$12\% coverage on the top), as illustrated in \autoref{fig:perturbation_examples}.
These variants simulate the common distortions introduced when memes are reshared online.
Specifically, Gaussian blurring simulates image degradation from compression, border cropping mimics partial trimming during editing, and UI overlays imitate interface bars that appear when memes are shared as screenshots. 
We choose perturbation levels that preserve human recognizability while introducing enough distortion to affect OCR, visual matching, and image retrieval.

\subsection{Baseline}
\label{sec:baseline}
Previous studies on meme understanding have primarily relied on VLMs~\cite{MSQYBZZ25}. Following this line, we use the vision–language model GPT-4o as our baseline, which adopts the same report-generation prompt as MemeTrace but without any external evidence. It directly generates the structured meme report from the input meme image. \new{We additionally include a stronger search-augmented baseline, GPT-4o with native web search, which shares the same backbone and prompt but relies solely on built-in text-based search instead of MemeTrace's retrieval modules.} To ensure consistent and reproducible outputs, we set the temperature to zero \new{for both}.

\subsection{Human Evaluation}
\label{sec:human_eval}
To evaluate the quality of generated provenance reports, we conduct an in-depth \new{blinded} human evaluation with three in-domain experts from the research team.
Each report's 11 fields are independently rated on two dimensions: \textbf{Informativeness}, which measures the completeness of the provided information, and \textbf{Correctness}, which measures its factual accuracy.
All ratings use a five-point Likert scale, and final scores are obtained by averaging across annotators. 
The main experiments are evaluated on \textbf{Meme100}, while ablation and robustness experiments use the \textbf{ablation subset} and \textbf{robustness subsets}.
Considering 11 fields and 2 dimensions, this configuration results in 7,920 rating items per annotator, yielding a total of 23,760 human-assigned scores across the three annotators.
Overall, the inter-annotator agreement is acceptable for the evaluation task: Intraclass Correlation Coefficient (ICC)~\citep{SF79} ranges from 0.502 to 0.600, which is reasonable given that three annotators rate open-ended report quality on a five-point Likert scale (Appendix~\ref{appendix:iaa}).
\new{We further verify these ratings with a blinded re-evaluation by two external annotators who are not co-authors, which also covers the search-augmented baseline (Appendix~\ref{appendix:blinded}).}

\subsection{Automatic Report Quality Metrics}
\label{sec:autometrics}
To enable scalable evaluation of provenance reports, we propose a suite of automatic report quality metrics to measure the quality of generated reports. These metrics are directly computed from the reports' textual content across three dimensions:  

\mypara{Numerical Specificity}
This dimension measures the amount of concrete factual detail included in a report. We use three metrics: 
\textit{digits\_per100w} (frequency of numerals per 100 words), 
\textit{years\_per100w} (frequency of 4-digit years per 100 words), and 
\textit{dates\_per100w} (frequency of explicit calendar dates per 100 words). 
Higher values reflect a more comprehensive grasp of factual details and temporal information in the background context.

\mypara{Entity Richness} 
This dimension evaluates the diversity of named entities in the reports. We compute the amount of entity for several categories, detected with spaCy~\cite{spacy} from three core fields \textit{about}, \textit{origin}, and \textit{spread}: 
\textit{ner\_person\_unique} (persons), 
\textit{ner\_org\_unique} (organizations), 
\textit{ner\_gpe\_unique} (countries), 
\textit{ner\_event\_unique} (events), 
\textit{ner\_loc\_unique} (locations), and 
\textit{ner\_fac\_unique} (facilities). 
Higher values in these metrics reflect a richer and more detailed understanding of entities and background information.

\mypara{Certainty and Context}
This dimension reflects how confident the agent is in the information it reports.
We measure the proportion of hedge expressions (\textit{hedge\_per100w}, e.g., ``maybe'', ``likely'') and the frequency of social-platform mentions (\textit{platforms\_per100w}, e.g., ``Twitter'', ``Reddit'').
Frequent hedge terms indicate lower agent confidence.
Similarly, when uncertain about the source, the agent tends to list multiple possible platforms.

\section{Results}
\subsection{Main Results}
We present the main experimental results of MemeTrace on Meme100, using both human evaluation and automated metrics.

\begin{table*}[t]
\centering
\caption{
Per-field human evaluation on Meme100, averaged across three annotators.
Abbreviations: str.=structure, ori.=origin, cul.=culture, aud.=target audience,
emo.=emotional impact, cog.=cognitive load, per.=persuasive technique.
}
\label{tab:human_eval_fields}
\setlength{\tabcolsep}{3.5pt}
\scalebox{0.85}{%
\begin{tabular}{@{}ll*{11}{c}@{}}
\toprule
\textbf{Metric} &
\textbf{System} &
\textbf{about} &
\textbf{str.} &
\textbf{ori.} &
\textbf{spread} &
\textbf{cul.} &
\textbf{usage} &
\textbf{aud.} &
\textbf{emo.} &
\textbf{cog.} &
\textbf{per.} &
\textbf{safe.} \\
\midrule
\multirow{2}{*}{\textit{Informativeness}}
& Baseline &
3.407 & 4.493 & 3.117 & 3.783 & 4.147 & 4.077 & 4.303 & 4.153 & 4.117 & 4.473 & 4.483 \\
& MemeTrace &
\textbf{4.917} & \textbf{4.927} & \textbf{4.533} & \textbf{4.670} & \textbf{4.883} &
\textbf{4.907} & \textbf{4.907} & \textbf{4.867} & \textbf{4.920} &
\textbf{4.913} & \textbf{4.777} \\
\midrule
\multirow{2}{*}{\textit{Correctness}}
& Baseline &
3.863 & 4.580 & 3.227 & 3.930 & 4.350 & 4.307 & 4.463 & 4.297 & 4.330 & 4.513 & 4.400 \\
& MemeTrace &
\textbf{4.880} & \textbf{4.957} & \textbf{4.553} & \textbf{4.733} &
\textbf{4.903} & \textbf{4.913} & \textbf{4.927} & \textbf{4.917} &
\textbf{4.903} & \textbf{4.917} & \textbf{4.670} \\
\bottomrule
\end{tabular}%
}
\end{table*}

\mypara{Human Evaluation}
We report the overall informativeness and correctness scores, averaged across the three annotators, in \autoref{tab:human_eval_overall}.
We also compute an overall \textit{adjusted\_score} as the mean of informativeness and correctness.
Compared with the VLM-only baseline, MemeTrace achieves consistently higher human ratings.
The informativeness of MemeTrace's reports increases from 4.050 to 4.838, correctness from 4.205 to 4.843, and the adjusted score from 4.128 to 4.841 (representing an overall improvement of 17\%).
\new{All three gaps are significant ($p<.001$).}
These results demonstrate the agent’s ability to produce detailed, factually accurate meme analyses.

\autoref{tab:human_eval_fields} presents the detailed human evaluation results for each field.
Compared to the baseline, MemeTrace achieves the highest scores across all evaluation fields, with particularly strong performance in fields \textit{about}, \textit{origin}, and \textit{spread}.
These results indicate that the agent shows superior capability in interpreting meme meaning, tracing its origin, and describing its spread across platforms, often producing more complete event timelines. 
More importantly, its performance in safety assessment (as indicated by \textit{safety\_flag}) also improves compared with the VLM baseline.
This brings greater practical value in real-world applications such as content moderation.
We further present case studies to illustrate its performance across different fields, in Appendix~\ref{sec: case_studies}.

\begin{table}[t]
\centering
\caption{
Overall human evaluation comparison between MemeTrace and baseline on Meme100.
Info.=Informativeness, Corr.=Correctness.
}
\label{tab:human_eval_overall}
\scalebox{0.85}{%
\renewcommand{\arraystretch}{1.1}
\setlength{\tabcolsep}{5pt}
\begin{tabular}{lccc}
\toprule
\textbf{System} & \textbf{Info.} & \textbf{Corr.} & \textbf{Adjusted} \\
\midrule
Baseline & 4.050 & 4.205 & 4.128 \\
\textbf{MemeTrace} & \textbf{4.838} & \textbf{4.843} & \textbf{4.841} \\
\bottomrule
\end{tabular}%
}
\end{table}

\mypara{Automatic Evaluation}
As shown in \autoref{tab:auto_metrics_agent_baseline}, the automatic evaluation results show that MemeTrace significantly outperforms the baseline across all metrics. 
On average, MemeTrace uses 2.088 numbers, 0.880 year mentions, and 0.380 explicit dates per 100 words, compared with 0.631, 0.168, and 0.020 for the baseline.
This suggests that MemeTrace provides more specific timelines and factual context when describing a meme’s origin and spread.
For platform mentions, the frequency decreases from 1.162 in the baseline to 0.653 per 100 words.
This is because baseline often lists several possible platforms when it lacks concrete evidence, while MemeTrace mentions platforms more selectively and ties them to specific evidence about the meme’s origin or spread.
For uncertainty expressions, the proportion of hedging phrases decreases from 1.42 in the baseline to 0.64 in MemeTrace, reflecting fewer speculative statements and more evidence-grounded claims. 
MemeTrace also mentions more unique entities across all categories, including persons (3.22 vs. 1.96), organizations (2.72 vs. 1.49), and events (0.11 vs. 0.01), which suggests richer contextual coverage.

\begin{table}[t]
\centering
\caption{
Automatic metric comparison between MemeTrace and baseline on Meme100.
}
\label{tab:auto_metrics_agent_baseline}
\scalebox{0.85}{%
\renewcommand{\arraystretch}{1.1}
\setlength{\tabcolsep}{4.2pt}
\begin{tabular}{lccc}
\toprule
\textbf{Metric} & \textbf{MemeTrace} & \textbf{Baseline} & $\Delta$ \\
\midrule
\texttt{digits\_per100w} & 2.088 & 0.631 & +1.457 \\
\texttt{years\_per100w}  & 0.880 & 0.168 & +0.712 \\
\texttt{dates\_per100w}  & 0.380 & 0.020 & +0.360 \\
\texttt{ner\_person\_unique} & 3.220 & 1.960 & +1.260 \\
\texttt{ner\_org\_unique}   & 2.720 & 1.490 & +1.230 \\
\texttt{ner\_gpe\_unique}   & 1.830 & 1.290 & +0.540 \\
\texttt{ner\_event\_unique} & 0.110 & 0.010 & +0.100 \\
\texttt{ner\_loc\_unique}   & 0.110 & 0.050 & +0.060 \\
\texttt{ner\_fac\_unique}   & 0.130 & 0.020 & +0.110 \\
\texttt{hedge\_per100w}  & 0.640 & 1.420 & --0.780 \\
\texttt{platforms\_per100w} & 0.653 & 1.162 & --0.509 \\
\bottomrule
\end{tabular}%
}
\end{table}

We compute Pearson's $r$ and Spearman's $\rho$ between each automatic metric and the human evaluation scores to assess their alignment.
The strongest correlations reach Pearson $r=0.78$ and Spearman $\rho=0.94$, indicating that these metrics capture quality signals consistent with human judgments and can serve as reliable proxies for scalable evaluation (see Appendix~\ref{sec:alignment} for details).

\subsection{Ablation Studies}
\mypara{Module Ablations}
To assess the contribution of each module, we individually disable the Local Database Retrieval, Reverse Image Search (RIS), and Open-Web Retrieval (Tavily) modules and evaluate their performance on the ablation subset. 
As shown in~\autoref{tab:ablation_robustness_overall}, the full system performs best, with an adjusted score of 4.853.
Removing Tavily causes the largest drop to 4.227, highlighting the importance of open-web evidence.
Removing RIS and LocalDB also lowers performance to 4.317 and 4.388, respectively.
Field-level results in~\autoref{tab:ablation_fields} show the same trend, with Tavily contributing most to information-intensive fields such as \textit{about} and \textit{origin}, while disabling RIS weakens the acquisition of the information about meme propagation.

\mypara{Model Comparisons}
We compare different models as MemeTrace's reasoning LLM on the ablation subset. 
Two representative multimodal LLM variants, \textit{Gemini 1.5 Pro} and \textit{Qwen2.5-VL-72B}, are evaluated with all other agent settings held constant.
As shown in \autoref{tab:ablation_robustness_overall}, both variants maintain strong overall performance \new{well above the VLM-only baseline (3.961)}, although they remain below \new{the default GPT-4o-based MemeTrace}.
\new{We do not rank Gemini against Qwen, as $n{=}20$ cannot separate two so closely performing backbones.}

\begin{table}[t]
\centering
\caption{
Human evaluation on the 20-meme \textbf{Ablation and Robustness subsets}.
Info.=Informativeness, Corr.=Correctness, Adj.=Adjusted Score.
}
\label{tab:ablation_robustness_overall}
\renewcommand{\arraystretch}{1.1}
\setlength{\tabcolsep}{4pt}
\scalebox{0.85}{
\begin{tabular}{lccc}
\toprule
\textbf{System/Dataset} & \textbf{Info.} & \textbf{Corr.} & \textbf{Adj.} \\
\midrule
Baseline & 3.886 & 4.035 & 3.961 \\
\midrule
MemeTrace (full) & \textbf{4.859} & \textbf{4.847} & \textbf{4.853} \\
MemeTrace (Gemini) & 4.452 & 4.450 & 4.451 \\
MemeTrace (Qwen) & 4.591 & 4.559 & 4.575 \\
\midrule
\textit{w/o LocalDB} & 4.394 & 4.382 & 4.388 \\
\textit{w/o RIS} & 4.335 & 4.300 & 4.317 \\
\textit{w/o Tavily} & 4.215 & 4.239 & 4.227 \\
\midrule
\textit{Robu\_Gauss\_Blur} & 4.292 & 4.202 & 4.247 \\
\textit{Robu\_UI\_Overlay} & 4.320 & 4.274 & 4.297 \\
\textit{Robu\_Border\_Crop} & 4.277 & 4.180 & 4.229 \\
\bottomrule
\end{tabular}
}
\end{table}

\subsection{Robustness Results}
\label{sec:robustness-result}

As shown in \autoref{tab:ablation_robustness_overall}, MemeTrace remains highly stable under three types of visual perturbations.
The system achieves an adjusted score of 4.853, while the scores on three robustness subsets show a decrease of about 0.56–0.62, roughly 12\%. 
In these subsets, UI overlay causes the smallest drop, showing that the RIS module remains effective even when parts of the interface are occluded.
Gaussian blur primarily reduces correctness because reduced text clarity hinders accurate recognition. 
Border cropping reduces both informativeness and correctness due to a lack of information near image boundaries.

Field-level results in Table~\ref{tab:ablation_fields} further show that the key fields \textit{about}, \textit{origin}, and \textit{structure} remain relatively stable on robustness subsets, suggesting that the system can reliably identify meme content and trace its origin even under visual distortions.

\subsection{Safety-Flag Analysis}
Existing harmful meme detection benchmarks mainly capture surface-level harmfulness, where decisions often require limited external context.
To evaluate safety assessment on emerging, context-dependent memes, we analyze the \textit{safety\_flag} field on Meme100 using human evaluation~(\autoref{tab:human_eval_fields}).
MemeTrace outperforms the VLM-only baseline in both informativeness (4.777 vs. 4.483) and correctness (4.670 vs. 4.400), indicating richer and more accurate safety assessments.
We further analyze safety-flag patterns: disinformation is the most frequent risk, appearing in 44\% of reports, followed by sensitive content (16\%)~(\autoref{fig:safety_flag_prevalence}).
Political and geopolitical memes show the highest disinformation rate ($\approx$50\%), while technology-related memes are rarely flagged as sensitive~(\autoref{fig:topic_disinfo_sensitive_prevalence}).
The co-occurrence matrix further shows strong coupling between disinformation and fake misattribution, suggesting that misinformation often involves misreferenced sources~(\autoref{fig:safety_flag_cooccurrence_heatmap}).
Execution traces are provided in Appendix~\ref{appendix:trace_log}.

\section{Conclusion}
This paper introduced meme provenance as a new research task in meme study. Unlike traditional meme classification or interpretation tasks, it aims to produce structured provenance reports that reconstruct the origin, spread, and cultural context of memes, and designs an LLM-based agent for this task. 
MemeTrace achieves substantially higher human evaluation scores in both informativeness and correctness on Meme100, outperforming the VLM-only baseline.
The proposed automatic report quality metrics align strongly with human evaluation scores, enabling scalable evaluation. Ablation studies further reveal complementary contributions of the Tavily, RIS, and LocalDB modules, with Tavily contributing the most. 
MemeTrace demonstrates strong robustness under visual perturbations.
This study provides a foundation for meme provenance analysis and content moderation research.
An interactive demonstration is currently available at \url{http://memetrace.duckdns.org/}.

\mypara{Implications}
Our work has real-world implications.
First, MemeTrace can serve as a valuable data collection and annotation tool for the research community: its structured, multi-perspective reports turn raw memes into richly annotated research artifacts, which support downstream systematic studies such as meme evolution and meme diffusion process.
Second, by comprehensively examining multiple aspects of a meme, the agent can reveal deeper layers of risk associated with it.
It can help platform moderators detect harmful, manipulative, or misleading content by supplying the contextual and cultural background that traditional classifiers cannot capture.

\section*{Limitations}

Despite its strong performance in provenance analysis, MemeTrace still has several limitations. 
First, the MemeDB dataset is limited in size and cannot yet comprehensively cover the diversity of meme templates, especially newly emerging or non-English variants. 
Second, the system requires multiple rounds of LLM calls and online retrievals, which leads to relatively high costs. A detailed analysis of system efficiency, latency, and deployment cost is provided in Appendix~\ref{appendix:system_efficiency}.
Third, our current evaluation focuses mainly on natural perturbations, while systematic red-teaming tests have not yet been performed. 
Potential vulnerabilities remain, such as source pollution caused by search-engine-optimized or low-quality websites, and maliciously injected or templated content hidden in webpages. Although the domain scoring algorithm mitigates some of these risks, it cannot completely prevent information pollution. 
Automatic report quality metrics offer useful quantitative indicators, but they are still imperfect. 
For example, the frequency of dates or numeric mentions can only reflect information density rather than factual accuracy.

\section*{Ethical Considerations}
Our study relies entirely on publicly accessible data, for example, the Meme100 dataset from X/4chan/Reddit/Facebook/TikTok/YouTube/Instagram. The meme agent complies with the terms of service of the platforms, robots.txt, and rate limits. Because all evidence collected by the agent is public, it does not introduce additional privacy risks. Human evaluation was carried out under content warnings with the option to skip, and no crowd workers or sensitive annotator data were involved. Although logs may include offensive material, they are restricted to offline research use and are not disseminated online.

\section*{LLM Usage Considerations}
LLMs were used for editorial purposes in this manuscript and to assist with system development. All outputs were inspected by the authors to ensure accuracy.

\bibliography{normal_generated_py3}

\appendix

\section{Meme Provenance Report Fields}
\label{appendix:provenance_fields}
\autoref{table: meme_provenance_aspects} summarizes the 11 fields used in the meme provenance report, organized into five main categories: its \textbf{semantics} (a general description of what the meme conveys), \textbf{spreading dynamics} (the event or context it derives from and how it circulates and gains traction across platforms), \textbf{cultural context} (its symbolic meaning, patterns of use, and target audience), \textbf{psychological influence} (its cognitive/emotional influence to the audience), and \textbf{safety} (whether it remains harmless to the public or instead propagates hateful ideologies, conspiracy theories, or disinformation/misinformation).
Three of the fields (about, origin, spread) are inspired by the structure of meme encyclopedia entries on the platform KnowYourMeme, while the remaining fields extend this pattern to capture cultural context, practical uses, and potential safety risks.
Together, the 11 fields enable a comprehensive interpretation of a meme’s evolution, sociocultural impact, and potential risks.

\begin{table*}[t]
\centering
\caption{11 fields in the meme provenance report.}
\begin{tabular}{p{0.16\linewidth} p{0.22\linewidth} p{0.52\linewidth}}
\toprule
\textbf{Main Category} & \textbf{Fields} & \textbf{Description} \\
\midrule
\multirow{2}{\linewidth}{\centering \textbf{Semantics}}
& \textbf{About} & Describes what the meme is about, including background information and contextual analysis. \\ \cmidrule(lr){2-3}
& \textbf{Structure} & Details the meme's visual and textual format, including layout, template, and constituent elements. \\
\midrule
\multirow{2}{\linewidth}{\centering \textbf{Spreading Dynamics}}
& \textbf{Origin} & Specifies when, where, and how the meme likely originated. \\ \cmidrule(lr){2-3}
& \textbf{Spread} & Explains how the meme spread across platforms or online communities. \\
\midrule
\multirow{3}{\linewidth}{\centering \textbf{Cultural Context}}
& \textbf{Culture} & Identifies the subcultures involved and the cultural symbolism the meme represents. \\ \cmidrule(lr){2-3}
& \textbf{Usage} & Describes the pragmatic or communicative use of the meme, such as sarcasm, trolling, or critique. \\ \cmidrule(lr){2-3}
& \textbf{Target Audience} & Provides a characterization of the meme's intended audience or demographic group. \\
\midrule
\multirow{3}{\linewidth}{\centering \textbf{Psychological Influence}}
& \textbf{Emotional Impact} & Captures the emotional responses that the meme is likely to evoke (e.g., humor, anger, fear). \\ \cmidrule(lr){2-3}
& \textbf{Cognitive Load} & Outlines the prior knowledge or contextual understanding required to interpret the meme. \\ \cmidrule(lr){2-3}
& \textbf{Persuasive Technique} & Identifies rhetorical or narrative strategies used (e.g., irony, visual analogy, false equivalence). \\
\midrule
{\centering \textbf{Safety Assessment}\par}
& \textbf{Safety Flag} & Indicates potential safety risks from a fixed set of categories: abusive, hateful, misogynous, cyberbullying, disinformation, fake misattribution, propaganda, offensive language, derogatory, trolling, sarcasm, unethical, political, satire, or sensitive. Multiple flags may apply to a single meme. \\
\bottomrule
\end{tabular}
\label{table: meme_provenance_aspects}
\end{table*}

\section{MemeDB Coverage and Technical Details}
\label{appendix:memedb}

\begin{figure}[t]
  \centering
  \includegraphics[width=0.95\linewidth]{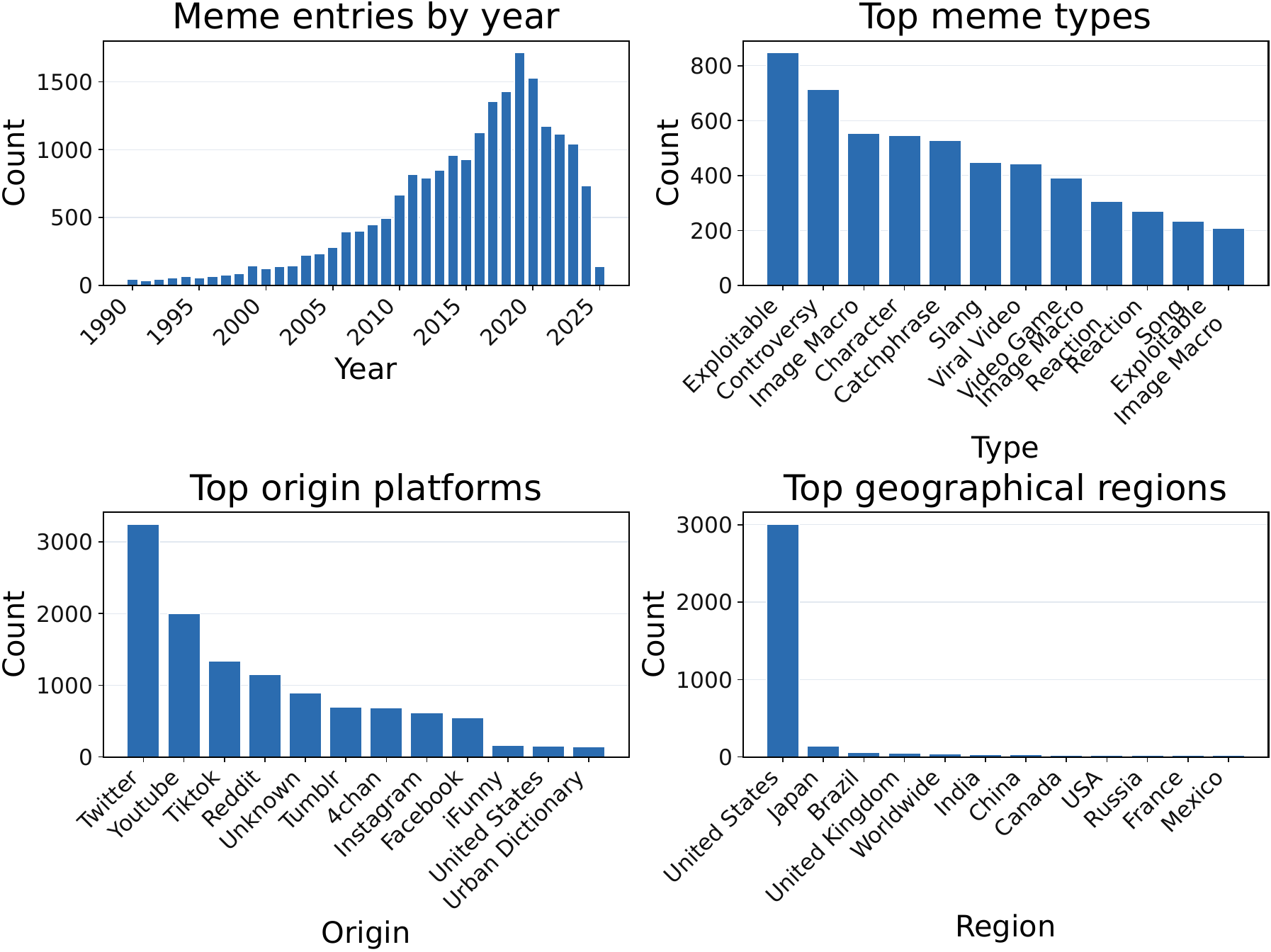}
  \caption{Coverage statistics of \textbf{MemeDB} (confirmed KYM entries, as of Jun.~2025):
  entries by year, types, origins, and regions.}
  \label{fig:memedb-coverage}
\end{figure}

We maintain a local meme database named \textbf{MemeDB}, which stores well-known memes that are already documented online.
This design enables fast retrieval and prevents repetitive search behaviors when similar memes appear in the future.
Our starting point for MemeDB is the documented memes from the KnowYourMeme website~\cite{kym}, the largest online meme encyclopedia.
Specifically, we collect all entries under the ``confirmed'' category, which contains 21,305 memes that have been manually reviewed by the KnowYourMeme editorial team, guaranteeing the reliability of the information.
Each entry in MemeDB contains a comprehensive set of structured metadata fields of the meme, including identifiers, textual descriptions, timestamps, origin, region, and tags.
\autoref{fig:memedb-coverage} presents the coverage statistics of \textbf{MemeDB}, which contains 21,305 confirmed entries from KnowYourMeme as of June 2025. 

After constructing the MemeDB dataset, we encode each meme’s image and its textual metadata using the OpenAI CLIP model~\cite{RKHRGASAMCKS21}, which projects both modalities into a shared embedding space. 
The resulting image and text embeddings are subsequently concatenated into a single vector, which is then used to build the FAISS index for efficient similarity search.
During the retrieval phase, the system queries this index with the input meme image and retrieves the three most relevant candidates.
Since CLIP similarity alone retrieves the nearest neighbors in MemeDB to the input meme, but not necessarily semantically related to it (e.g., different meanings despite identical visual features), we employ a multimodal LLM, GPT-4o, to assess the semantic relevance of each candidate and assign similarity scores between 0 and 10. 
Only candidates with scores above a fixed threshold of 8 are selected with their metadata as supporting evidence for subsequent provenance reasoning.
This ensures that only contextually relevant matches are passed to later modules. 

\section{Domain Scoring Algorithm Details}
\label{appendix:domain_scoring}

In the Reverse Image Search (RIS) branch, the system receives candidate URLs returned by Google RIS.
To avoid unnecessary resource consumption from crawling all candidates including unrelated ones, each URL is ranked according to the adjusted score of its domain, computed by a domain-based scoring algorithm, and only the top-5 URLs are selected for crawling.
Each domain’s adjusted score in the database reflects the estimated quality of its content.
After scraping, the RIS evidence checker assigns each URL a usefulness score, which evaluates the quality of the retrieved webpages.
These new usefulness scores are then used to update the corresponding domain’s adjusted score using the scoring algorithm, which is written back to the database.
The algorithm computes a confidence-bound-based lower bound on domain quality with a decaying exploration bonus to balance exploitation and exploration, prioritizing high-quality domains (e.g., \emph{knowyourmeme.com}) while giving new domains opportunities for exposure.

\section{RIS Evidence Checker Details}
\label{appendix:ris-example}

The crawled webpages often contain substantial noise, such as HTML markup and advertisements.
To address this, the crawled content is processed by the RIS evidence checker module, which uses a multimodal LLM (GPT-4o) to clean the extracted text and aggregate it into a structured summary with inline citations.
Furthermore, the checker assigns a usefulness score between 0 and 1 to each source, reflecting its relevance and evidential value, and subsequently uses it to update the dynamic domain-quality record within the domain-based scoring algorithm.
Since the information quality of external sources can vary over time, these continuously updated records enable the system to adapt accordingly.

\autoref{fig:example-ris} shows an example output from the RIS checker node.

\begin{figure}[t]
\centering
\begin{lstlisting}[basicstyle=\ttfamily\tiny, breaklines=true, showstringspaces=false]
{
  "summary": "Key Facts: 'The Last Supper' by Leonardo da Vinci was painted between 1495 and 1498 [S1]...",
  "sources": [
    "https://cenacolovinciano.org/en/museum/the-works/...",
  ],
  "scores": {
    "https://www.amazon.com/Last-Supper-Leonardo-daVinci-T-Shirt/dp/B0C5Y933ZN": 0.0,
    "https://cenacolovinciano.org/en/museum/the-works/...": 0.8,
  }
}
\end{lstlisting}
\caption{Example output from the RIS checker node, showing a structured summary with key facts, source attributions, and per-URL usefulness scores derived from reverse image search results.}
\label{fig:example-ris}
\end{figure}

\section{Tavily Quality Checker Details}
\label{appendix:tavily-checker}

Similar to the RIS evidence checker, the retrieved results from Tavily Search are processed by the Tavily quality checker module, a multimodal LLM that filters noise, preserves relevant details, and organizes them into a structured summary with inline citations.
The checker also assigns a quality score between 0 and 1 to assess the usefulness of the information.
Finally, the summary and its score are passed back to the evaluator to guide subsequent actions.

\section{MemeTrace Implementation Details}
\label{appendix:implementation}

MemeTrace is implemented using the LangGraph framework~\cite{LangGraph}, which is a low-level orchestration open source framework for building, managing, and deploying complex LLM agents.
To prevent infinite searches, MemeTrace sets a maximum iteration limit, with a default value of 5.
The main experiments use GPT-4o with temperature set to zero, accessed via the OpenAI API~\cite{OpenAIAPI}, to ensure deterministic behavior. 
For model comparison experiments, we replace the backend model with Gemini-Pro-1.5 and Qwen2.5-VL-72B-Instruct via the OpenRouter API~\cite{openrouter}, keeping all other components and configurations fixed, including the same iteration limit, temperature, and prompt templates, to ensure comparability across systems.

The system stores domain scores and MemeDB's metadata in a SQLite database, and saves execution logs as Markdown files to enable auditing. 
All experiments are conducted on a MacBook Pro (M4 Max, 36 GB RAM).

\section{Meme100 Dataset Construction}
\label{appendix:meme100}

\begin{figure}[t]
  \centering
  \includegraphics[width=\linewidth]{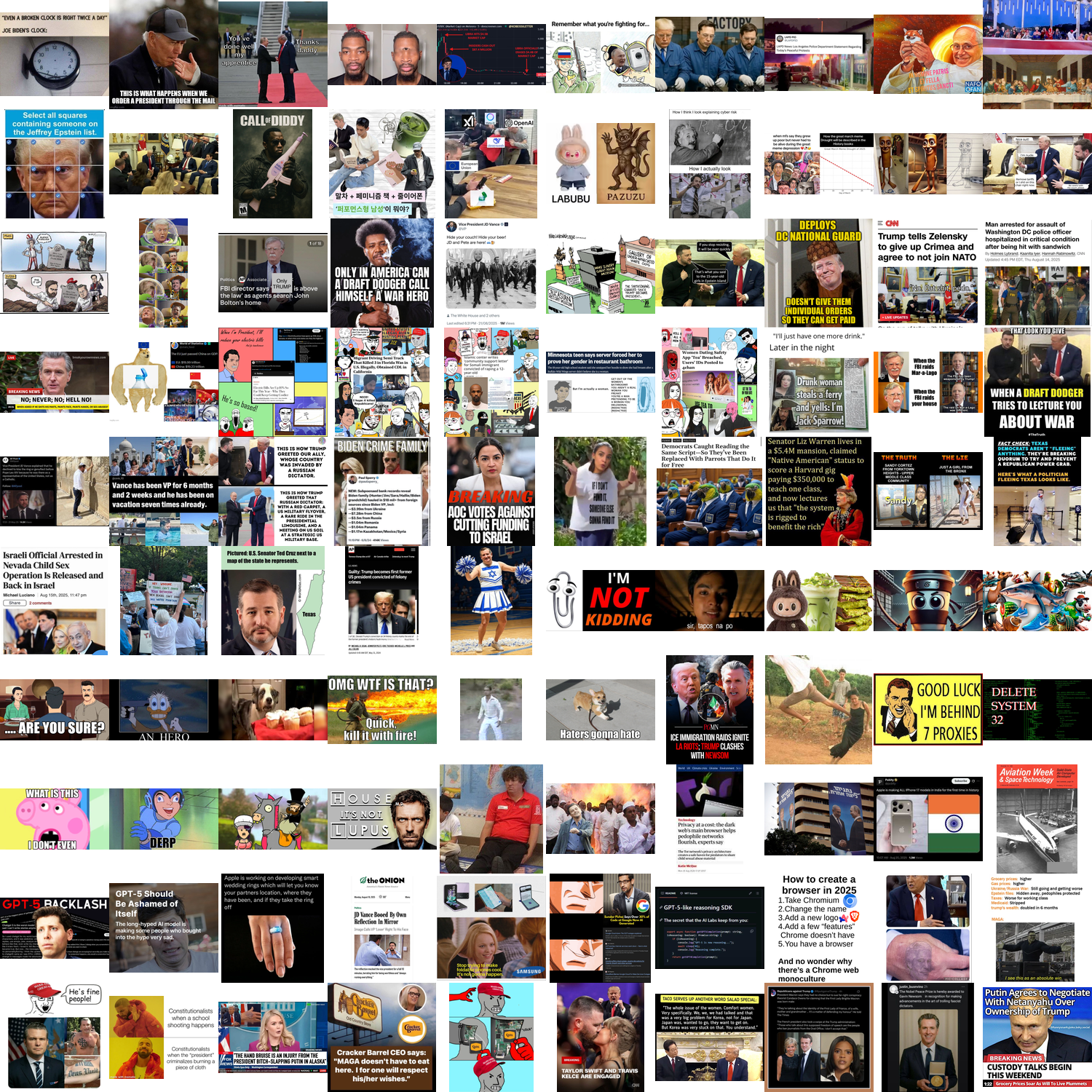}
  \caption{Overview of all 100 memes in the Meme100 dataset.}
  \label{fig:meme100-overview}
\end{figure}

\begin{figure}[t]
  \centering

  \begin{minipage}[t]{0.48\linewidth}
    \centering
    \includegraphics[width=\linewidth]{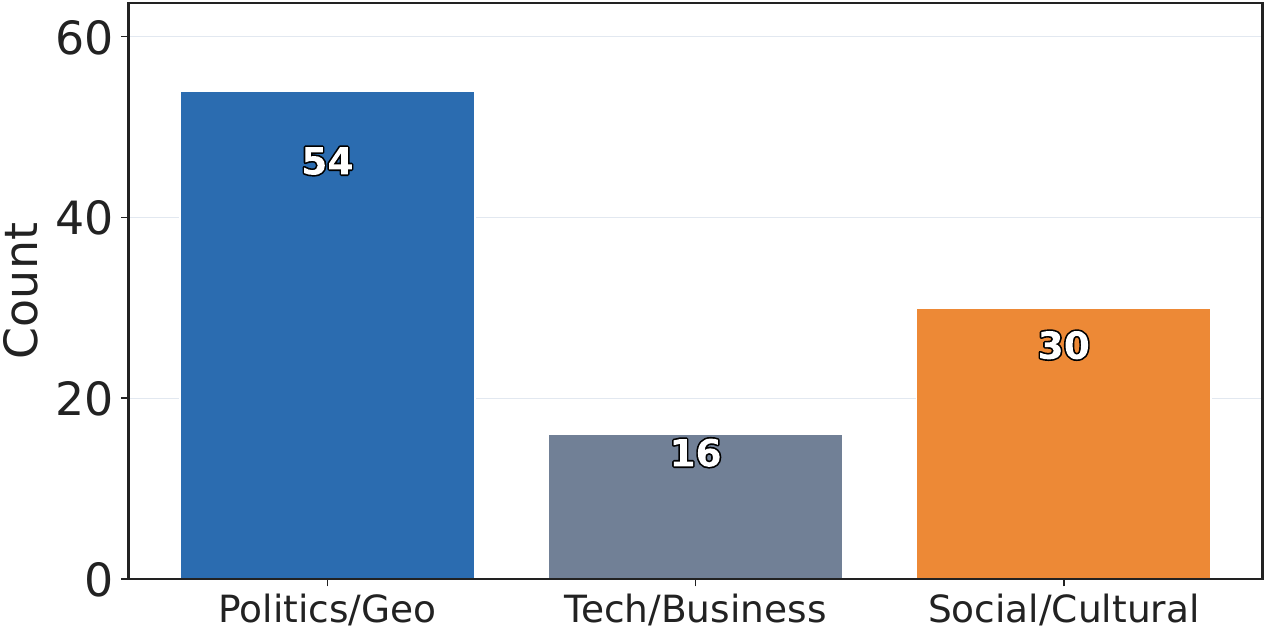}
    {\scriptsize \mbox{(a) Topic Distribution}}
  \end{minipage}
  \hfill
  \begin{minipage}[t]{0.48\linewidth}
    \centering
    \includegraphics[width=\linewidth]{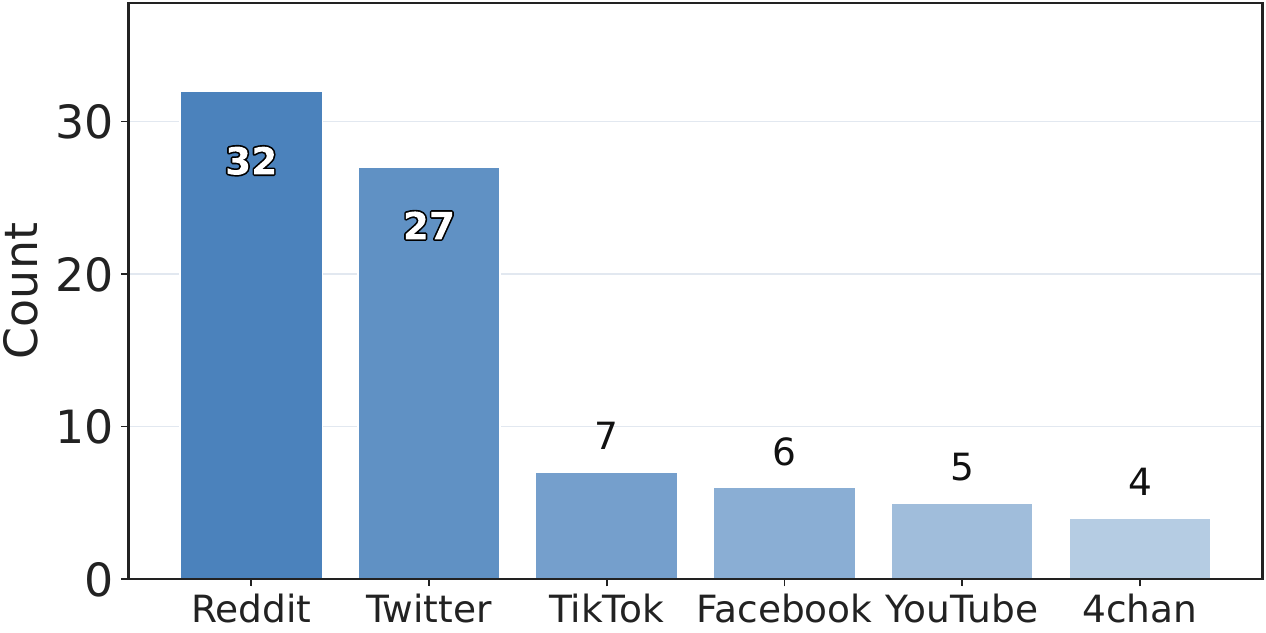}
    {\scriptsize \mbox{(b) Platform Distribution}}
  \end{minipage}

  \begin{minipage}[t]{0.48\linewidth}
    \centering
    \includegraphics[width=\linewidth]{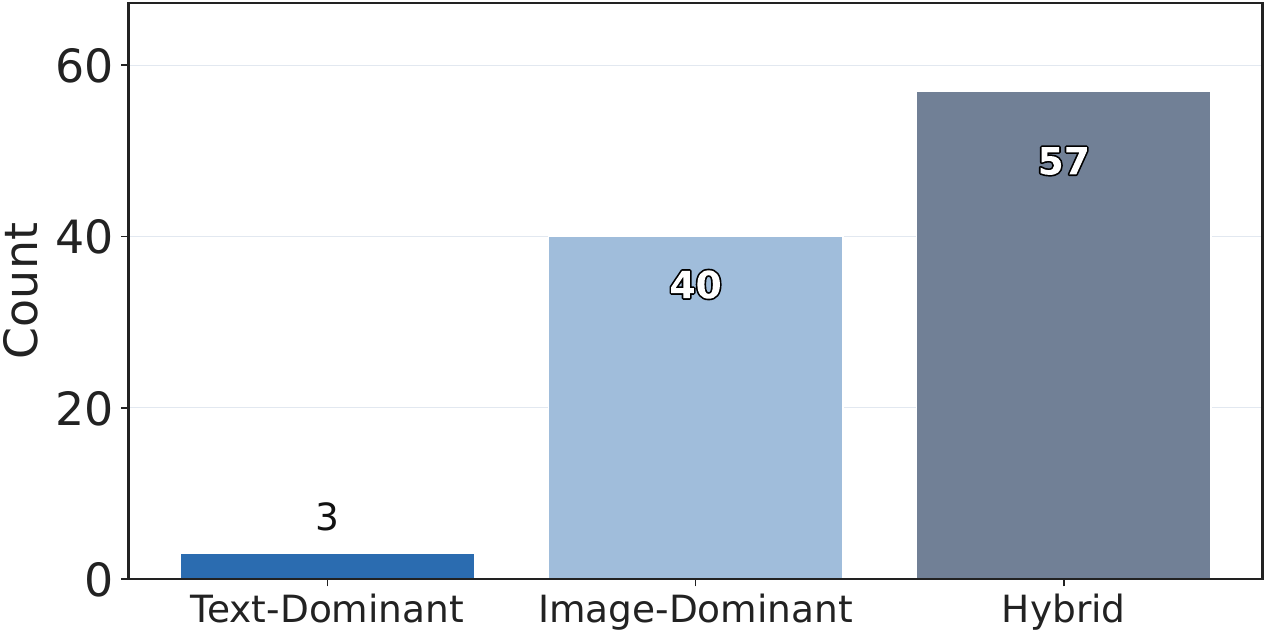}
    {\scriptsize \mbox{(c) Modality Distribution}}
  \end{minipage}
  \hfill
  \begin{minipage}[t]{0.48\linewidth}
    \centering
    \includegraphics[width=\linewidth]{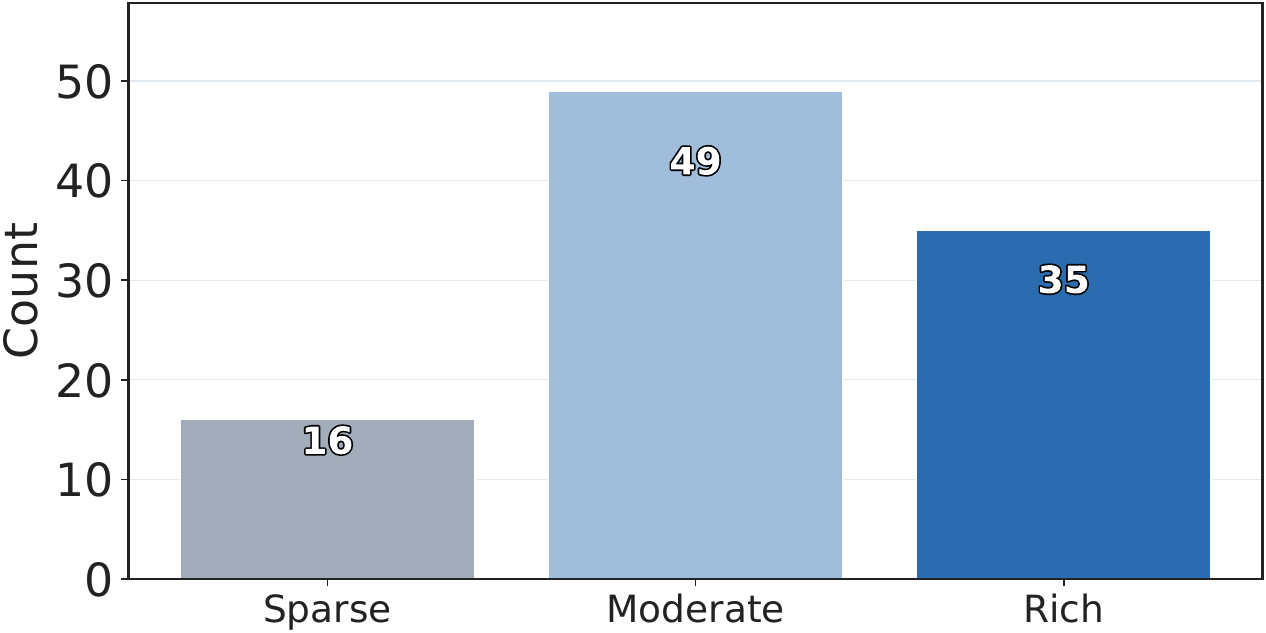}
    {\scriptsize \mbox{(d) Sparsity Distribution}}
  \end{minipage}

  \caption{
  Distributions of \textbf{Meme100}: (a) Topic, (b) Platform, (c) Modality, (d) Sparsity.
  }
  \label{fig:meme100-dist}
\end{figure}

We construct a curated evaluation dataset consisting of 100 memes collected from diverse social platforms. We adopt an event-driven collection strategy to capture memes that are actively used in high-impact public discourse. 
Candidate memes were retrieved by querying major social platforms (Reddit, Twitter/X, 4chan, Facebook, TikTok, YouTube, and Instagram) using keywords associated with recent high-salience events, followed by filtering posts with engagement exceeding 5,000 interactions. 
The selected events span multiple categories, including politics and geopolitics, social and cultural trends, and technology-related controversies. Examples include the 2024 U.S. presidential election, the Russia–Ukraine conflict, and large-scale protests in Los Angeles in mid-2025. 
To better cover diverse meme derivations and enhance dataset representativeness, we additionally include a subset of classic meme templates sourced from the Know Your Meme encyclopedia~\cite{kym}, focusing on widely circulated templates that emerged more than 10 years ago.
In contrast, newly emerged memes correspond to event-driven variants or templates that gained prominence within the past year.
\autoref{fig:meme100-overview} visualizes all 100 memes in Meme100, and \autoref{fig:meme100-dist} presents the overall distribution statistics.
Although limited in size, Meme100 focuses on representative and reusable meme templates, which are sufficient to capture the dominant provenance patterns observed in real-world discourse.

\paragraph{Data Sources and Licensing}
Memes were collected from publicly accessible social media platforms (Reddit, Twitter/X, 4chan, Facebook, TikTok, YouTube, Instagram) and Know Your Meme~\cite{kym}, subject to their respective terms of service.
All content was publicly posted and is used for academic research under fair use.
The dataset will be released under CC BY-NC 4.0 license.

\section{Alignment Between Automatic and Human Evaluation}
\label{sec:alignment}
\begin{figure*}[t]
\centering
\includegraphics[width=0.49\linewidth]{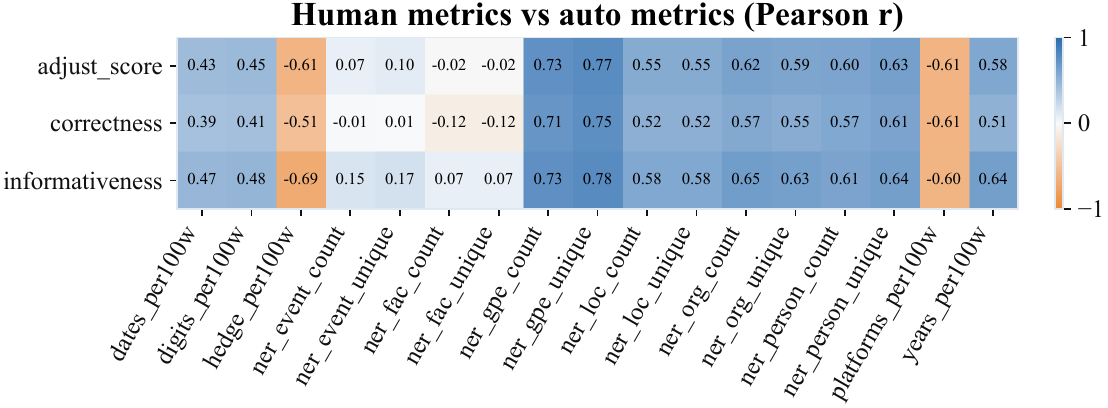}
\hfill
\includegraphics[width=0.49\linewidth]{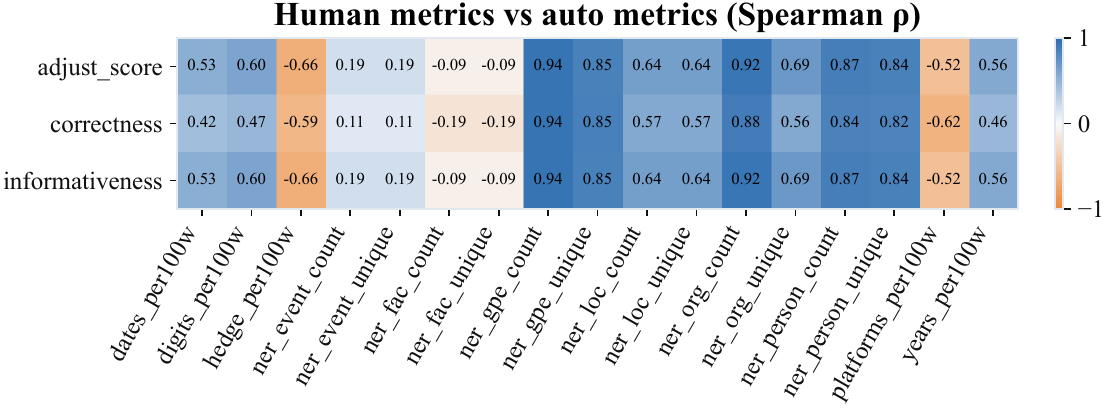}
\caption{Alignment between human evaluation and automatic metrics: (left) Pearson $r$, (right) Spearman $\rho$.}
\label{fig:heatmap_alignment}
\end{figure*}

To examine the alignment between automatic metrics and human evaluation, we computed both the Pearson correlation coefficient (r) and the Spearman rank correlation ($\rho$) between human evaluation scores for informativeness, correctness, and adjusted score and all automatic metrics. 
The heatmaps in  ~\autoref{fig:heatmap_alignment} display both Pearson and Spearman correlation results.
Entity-based metrics show the strongest alignment with human evaluation, particularly \texttt{ner\_gpe\_unique}, \texttt{ner\_org\_unique}, and their count variants, where unique measures distinct entities and count represents total entity mentions.
Temporal and numerical metrics, such as \texttt{dates\_per100w}, \texttt{digits\_per100w}, and \texttt{years\_per100w}, show moderate positive correlations, indicating that the richness of specific temporal or quantitative information can also reflect the quality of the report well.
In contrast, hedging frequency and platform enumeration are negatively correlated with human evaluation scores, which is consistent with our expectations, as excessive hedging or platform mentions often reflect uncertain or speculative reasoning, which are always validated by human annotators as low-quality reports.
Overall, the results show that automatic metrics can serve as reliable metrics enabling scalable assessment of report quality.

\section{Inter-Annotator Agreement}
\label{appendix:iaa}
Tables~\ref{tab:iaa_variance} and~\ref{tab:iaa_alpha_rho} provide detailed inter-annotator agreement statistics. The main experiment shows low variance (0.108 overall), indicating high consistency among annotators.
ICC=0.502 (moderate reliability) further confirms acceptable agreement for this open-ended expert evaluation task.

\begin{table}[t]
\centering
\caption{Inter-annotator agreement measured by mean rating variance (lower is better).}
\label{tab:iaa_variance}
\scalebox{0.78}{%
\renewcommand{\arraystretch}{1.10}
\setlength{\tabcolsep}{4pt}
\begin{tabular}{lccc}
\toprule
\textbf{System} & \textbf{Var(info)} & \textbf{Var(corr)} & \textbf{Var(overall)} \\
\midrule
Baseline              & 0.6191 & 0.8285 & 0.6025 \\
MemeTrace (full)     & 0.1448 & 0.1630 & 0.1076 \\
MemeTrace (Gemini)    & 0.8470 & 0.8833 & 0.8148 \\
MemeTrace (Qwen)       & 0.6636 & 0.7530 & 0.6663 \\
w/o LocalDB      & 0.9318 & 1.0197 & 0.8992 \\
w/o RIS          & 1.2242 & 1.3348 & 1.2049 \\
w/o Tavily       & 1.4727 & 1.6091 & 1.4739 \\
Robu\_Gauss\_Blur   & 1.1970 & 1.2576 & 1.1242 \\
Robu\_UI\_Overlay  & 1.1379 & 1.1288 & 1.0530 \\
Robu\_Border\_Crop    & 1.2636 & 1.2212 & 1.0992 \\
\bottomrule
\end{tabular}%
}
\end{table}

\begin{table}[t]
\centering
\caption{Inter-annotator agreement measured by mean rating variance and ICC.}
\label{tab:iaa_alpha_rho}
\scalebox{0.9}{%
\renewcommand{\arraystretch}{1.10}
\setlength{\tabcolsep}{6pt}
\begin{tabular}{lcc}
\toprule
\textbf{System} & \textbf{Var(overall)} & \textbf{ICC(overall)} \\
\midrule
Baseline         & 0.6025 & 0.6004 \\
MemeTrace (full) & 0.1076 & 0.5021 \\
\bottomrule
\end{tabular}%
}
\end{table}

\new{\section{External Blinded Re-Evaluation and Search-Augmented Baseline}}
\label{appendix:blinded}

\new{\mypara{Protocol}
Two external annotators who volunteered for this study, neither of whom is a co-author, independently re-scored three systems on the ablation subset, with system identifiers removed and report order randomized: MemeTrace, the VLM-only baseline, and the search-augmented baseline (GPT-4o with native web search).}

\new{\mypara{Agreement}
Agreement between the two annotators is strong: pooled over the three systems, ICC(C,k)$\,=\,$0.801.}

\new{\mypara{Results}
\autoref{tab:blinded_overall} shows MemeTrace outperforms both baselines.
The search-augmented baseline substantially outperforms the plain baseline (4.594 vs.\ 3.964), confirming that external evidence drives provenance quality; yet MemeTrace wins on 18 of 20 memes, demonstrating that its gains stem from its specialized retrieval modules rather than generic web search alone.}

\begin{table}[t]
\centering
\caption{\new{Overall blinded human evaluation on the 20-meme ablation subset, averaged over the two external annotators.
WebSearch = GPT-4o with native web search.
Info.=Informativeness, Corr.=Correctness.}}
\label{tab:blinded_overall}
\scalebox{0.85}{%
\renewcommand{\arraystretch}{1.1}
\setlength{\tabcolsep}{5pt}
\begin{tabular}{lccc}
\toprule
\textbf{System} & \textbf{Info.} & \textbf{Corr.} & \textbf{Adjusted} \\
\midrule
Baseline & 3.877 & 4.050 & 3.964 \\
WebSearch & 4.607 & 4.582 & 4.594 \\
\textbf{MemeTrace} & \textbf{4.800} & \textbf{4.780} & \textbf{4.790} \\
\bottomrule
\end{tabular}%
}
\end{table}

\section{Per-Field Human Evaluation Results}
Table~\ref{tab:ablation_fields} provides detailed per-field human evaluation scores.

\begin{table*}[t]
\centering
\caption{
Per-field human evaluation under module ablations, model variants, and robustness perturbations.
Abbreviations: aud.=target audience, emo.=emotional impact,
cog.=cognitive load, pers.=persuasive technique, Robu=Robustness, Gauss=Gaussian, Crop=Cropping.
}
\label{tab:ablation_fields}
\scalebox{0.85}{%
\renewcommand{\arraystretch}{1.1}
\setlength{\tabcolsep}{5.5pt}
\begin{tabular}{lccccccccccc}
\toprule
\textbf{System} &
\textbf{about} & \textbf{structure} & \textbf{origin} & \textbf{spread} &
\textbf{culture} & \textbf{usage} & \textbf{aud.} &
\textbf{emo.} & \textbf{cog.} &
\textbf{pers.} & \textbf{safety} \\
\midrule
\multicolumn{12}{l}{\textit{Informativeness}} \\
Baseline &
3.133 & 4.500 & 2.950 & 3.850 & 3.900 & 4.000 & 4.200 & 4.000 & 3.767 & 4.183 & 4.267 \\
MemeTrace (full) &
\textbf{4.967} & \textbf{4.917} & \textbf{4.733} & \textbf{4.700} & \textbf{4.817} & \textbf{4.950} & \textbf{4.883} & \textbf{4.833} & \textbf{4.983} & \textbf{4.900} & \textbf{4.767} \\
MemeTrace (Gemini) &
4.950 & 4.417 & 4.650 & 4.167 & 4.467 & 4.400 & 4.683 & 4.083 & 4.433 & 4.200 & 4.517 \\
MemeTrace (Qwen) &
4.983 & 4.700 & 4.583 & 4.317 & 4.550 & 4.667 & 4.700 & 4.450 & 4.683 & 4.383 & 4.483 \\
w/o LocalDB &
4.783 & 4.450 & 4.450 & 4.117 & 4.367 & 4.300 & 4.417 & 4.217 & 4.467 & 4.250 & 4.517 \\
w/o RIS &
4.800 & 4.483 & 4.367 & 4.000 & 4.283 & 4.250 & 4.417 & 4.150 & 4.283 & 4.150 & 4.500 \\
w/o Tavily &
4.517 & 4.267 & 4.067 & 3.950 & 4.283 & 4.083 & 4.400 & 4.033 & 4.317 & 3.983 & 4.467 \\
Robu\_Gauss\_Blur &
4.633 & 4.367 & 4.083 & 4.117 & 4.217 & 4.250 & 4.433 & 4.183 & 4.400 & 4.133 & 4.400 \\
Robu\_UI\_Overlay &
4.700 & 4.383 & 4.267 & 4.033 & 4.317 & 4.167 & 4.383 & 4.150 & 4.433 & 4.117 & 4.567 \\
Robu\_Border\_Crop &
4.683 & 4.250 & 4.300 & 3.933 & 4.183 & 4.117 & 4.367 & 4.200 & 4.417 & 4.133 & 4.467 \\
\midrule
\multicolumn{12}{l}{\textit{Correctness}} \\
Baseline &
3.383 & 4.517 & 3.100 & 4.067 & 4.150 & 4.067 & 4.367 & 4.217 & 4.067 & 4.267 & 4.183 \\
MemeTrace (full) &
\textbf{4.867} & \textbf{4.967} & \textbf{4.633} & \textbf{4.700} & \textbf{4.917} & \textbf{4.917} & \textbf{4.950} & \textbf{4.900} & \textbf{4.933} & \textbf{4.933} & \textbf{4.600} \\
MemeTrace (Gemini) &
4.783 & 4.467 & 4.633 & 4.333 & 4.483 & 4.450 & 4.683 & 4.150 & 4.417 & 4.267 & 4.283 \\
MemeTrace (Qwen) &
4.967 & 4.683 & 4.667 & 4.450 & 4.550 & 4.600 & 4.667 & 4.433 & 4.717 & 4.383 & 4.033 \\
w/o LocalDB &
4.700 & 4.417 & 4.417 & 4.283 & 4.400 & 4.433 & 4.450 & 4.300 & 4.450 & 4.250 & 4.100 \\
w/o RIS &
4.667 & 4.500 & 4.333 & 4.067 & 4.283 & 4.300 & 4.417 & 4.233 & 4.283 & 4.083 & 4.133 \\
w/o Tavily &
4.633 & 4.283 & 4.200 & 4.133 & 4.350 & 4.167 & 4.400 & 4.083 & 4.350 & 4.050 & 3.983 \\
Robu\_Gauss\_Blur &
4.483 & 4.283 & 3.967 & 4.200 & 4.150 & 4.183 & 4.267 & 4.217 & 4.350 & 4.133 & 3.983 \\
Robu\_UI\_Overlay &
4.650 & 4.333 & 4.283 & 4.167 & 4.267 & 4.183 & 4.333 & 4.167 & 4.367 & 4.117 & 4.150 \\
Robu\_Border\_Crop &
4.467 & 4.200 & 4.067 & 4.017 & 4.117 & 4.067 & 4.300 & 4.200 & 4.333 & 4.167 & 4.050 \\
\bottomrule
\end{tabular}%
}
\end{table*}

\section{Qualitative Case Studies}
\label{sec: case_studies}
\begin{figure*}[t]
\centering
\setlength{\fboxsep}{6pt}
\fcolorbox{black}{gray!5}{
\begin{minipage}{0.94\linewidth}
  \centering
  \includegraphics[width=0.35\linewidth]{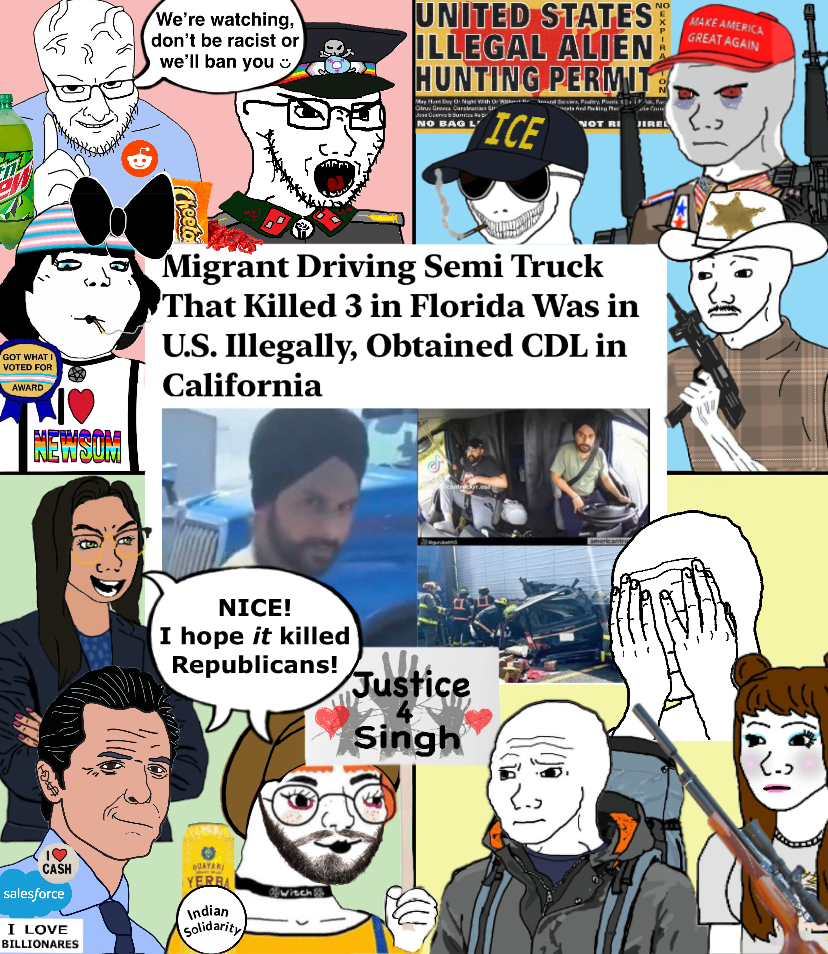}

  {\scriptsize
  \renewcommand{\arraystretch}{1.05}
  \begin{tabularx}{0.96\linewidth}{@{}X X@{}}
    \textbf{(a) MemeTrace (Agent)} &
    \textbf{(b) Baseline (VLM)} \\
    \begin{minipage}[t]{0.94\linewidth}
      \raggedright
      \textbf{About:} The meme is a satirical commentary on immigration, political polarization, and media representation in the U.S. It references a real incident involving Harjinder Singh, a 28-year-old illegal immigrant from India, who caused a fatal accident on the Florida Turnpike on Aug 12, 2025, resulting in three deaths. Singh, who entered the U.S. illegally in 2018, obtained a commercial driver's license in California and was charged with vehicular homicide. The collage mixes characters and symbols (ICE, MAGA, social-justice imagery) to critique both political sides and highlight tensions around immigration policies. Speech bubbles like ``NICE! I hope it killed Republicans!'' and ``Justice 4 Singh'' add shock value; ``We're watching, don't be racist or we'll ban you'' satirizes online moderation/censorship.\\[4pt]
      \textbf{Origin:} Likely from online communities focused on political satire (e.g., Reddit/4chan) where such collage formats are common.\\[4pt]
      \textbf{Spread:} Circulated on platforms with active political discourse (Twitter/X, Reddit, Facebook), resonating with audiences engaged in immigration and policy debates.\\[4pt]
      \textbf{Culture:} Belongs to political-satire subculture and internet humor around current events and social justice.\\[4pt]
      \textbf{Structure:} A collage of images and text with caricatures, speech bubbles, MAGA hat/ICE cap, and a headline about the incident.\\[4pt]
      \textbf{Usage:} Sarcasm and critique intended to provoke discussion on immigration, political identity, and social justice.\\[4pt]
      \textbf{Target Audience:} Politically engaged users interested in immigration issues and satirical commentary.\\[4pt]
      \textbf{Emotional Impact:} Mix of humor, anger, and shock depending on viewers' political stance.\\[4pt]
      \textbf{Cognitive Load:} Moderate; requires knowledge of the Singh case, U.S. immigration policy, and common political symbols.\\[4pt]
      \textbf{Persuasive Technique:} Irony and visual analogy to highlight perceived hypocrisies in immigration discourse.\\[4pt]
      \textbf{Safety Flag:} [\texttt{disinformation}, \texttt{sensitive}, \texttt{sarcasm}]
    \end{minipage}
    &
    \begin{minipage}[t]{0.94\linewidth}
      \raggedright
      \textbf{About:} A collage of internet characters and symbols used in political satire. It references a headline about a migrant driving a semi-truck that killed three people in Florida, touching on illegal immigration and state policies. Characters include Wojak/NPCs representing ideologies. There are cues about California (Gov. Gavin Newsom, tech companies) suggesting a liberal bias. Signs like ``Justice 4 Singh'' and the line ``NICE! I hope it killed Republicans!'' imply a sarcastic/critical tone toward perceived liberal reactions.\\[4pt]
      \textbf{Origin:} Likely from 4chan/Reddit where political and satirical collages are common and formats are remixed.\\[4pt]
      \textbf{Spread:} Shared across Twitter/X, Facebook, and meme communities on Reddit.\\[4pt]
      \textbf{Culture:} Internet meme culture—political satire and commentary subcultures.\\[4pt]
      \textbf{Structure:} Collage of images + headline text combined with recognizable meme characters.\\[4pt]
      \textbf{Usage:} Political critique and sarcasm aimed at provoking reactions across groups.\\[4pt]
      \textbf{Target Audience:} Politically active internet users familiar with meme tropes.\\[4pt]
      \textbf{Emotional Impact:} Anger, humor, or frustration depending on stance.\\[4pt]
      \textbf{Cognitive Load:} Moderate; needs familiarity with meme characters, political figures, and the news event.\\[4pt]
      \textbf{Persuasive Technique:} Irony and visual analogy to critique policies and ideological responses.\\[4pt]
      \textbf{Safety Flag:} [\texttt{sarcasm}, \texttt{trolling}, \texttt{sensitive}]
    \end{minipage}
  \end{tabularx}}
\end{minipage}}
\caption{
Qualitative comparison on the ``immigration/Florida–turnpike'' meme.
}
\label{fig:case_img33_immigration}
\end{figure*}

\begin{figure*}[t]
\centering
\setlength{\fboxsep}{6pt}
\fcolorbox{black}{gray!5}{
\begin{minipage}{0.94\linewidth}
  \centering
  \includegraphics[width=0.25\linewidth]{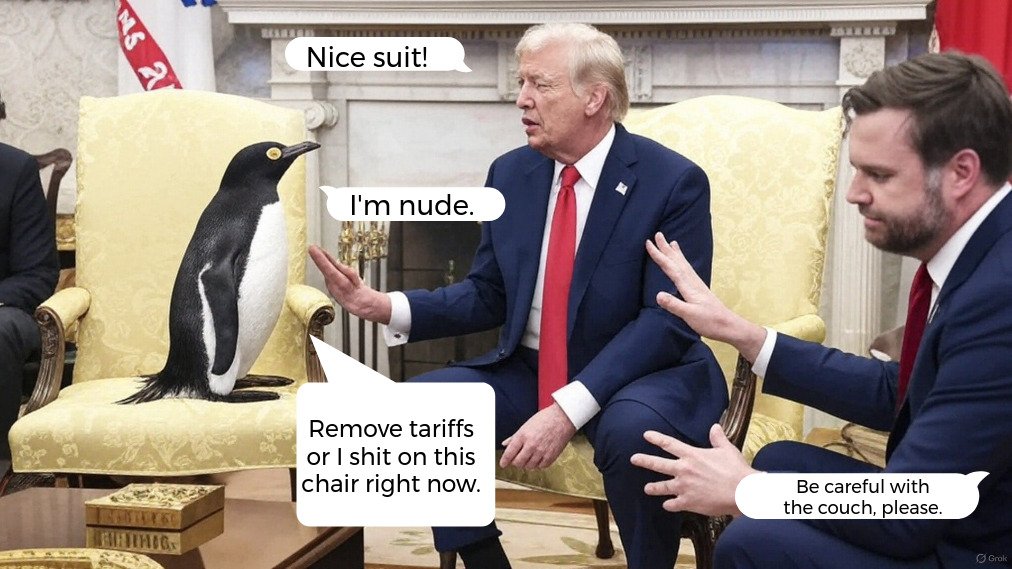}

  {\scriptsize
  \renewcommand{\arraystretch}{1.05}
  \begin{tabularx}{0.96\linewidth}{@{}X X@{}}
    \textbf{(a) MemeTrace (Agent)} &
    \textbf{(b) Baseline (VLM)} \\
    \begin{minipage}[t]{0.94\linewidth}
      \raggedright
      \textbf{About:} The meme features a penguin sitting in the Oval Office alongside Donald Trump and another individual, humorously discussing tariffs. The penguin, representing the wildlife of the Heard and McDonald Islands, is depicted as negotiating with Trump, threatening to ``shit on this chair'' if tariffs are not removed. This satirical image critiques the absurdity of Trump's Apr 2, 2025 decision to impose a 10\% tariff on exports from these uninhabited islands, which are home to penguins. The meme highlights the perceived irrationality of targeting such remote locations, using humor to underscore public confusion and criticism. The penguin's presence serves as a metaphor for the islands' non-human inhabitants, emphasizing the absurdity of the policy. The meme is part of a broader trend of penguin-themed memes that went viral on platforms like X (formerly Twitter) and Instagram, mocking the tariffs and their geopolitical implications.\\[4pt]
      \textbf{Origin:} The meme likely originated shortly after the Apr 2, 2025 announcement of tariffs on the Heard and McDonald Islands. It draws from a viral trend of penguin memes that emerged in response to the policy.\\[4pt]
      \textbf{Spread:} The meme spread widely on social media platforms such as X and Instagram, becoming part of a viral trend that included various penguin-themed memes critiquing the tariffs.\\[4pt]
      \textbf{Culture:} The meme belongs to internet meme culture, specifically within the realm of political satire and humor. It reflects a subculture that uses absurdity to critique political decisions.\\[4pt]
      \textbf{Structure:} The meme is a digitally altered image featuring a penguin in a formal setting, with speech bubbles conveying humorous dialogue.\\[4pt]
      \textbf{Usage:} The meme is used for satire and critique, employing humor to comment on the perceived absurdity of the tariffs.\\[4pt]
      \textbf{Target Audience:} Social media users interested in political satire, particularly those critical of Trump's trade policies.\\[4pt]
      \textbf{Emotional Impact:} The meme evokes humor and amusement, while also prompting reflection on the irrationality of certain political decisions.\\[4pt]
      \textbf{Cognitive Load:} Understanding the meme requires some prior knowledge of the tariff announcement and the geopolitical context involving the Heard and McDonald Islands.\\[4pt]
      \textbf{Persuasive Technique:} The meme uses irony and visual analogy, juxtaposing a penguin in a political setting to highlight the absurdity of the tariffs.\\[4pt]
      \textbf{Safety Flag:} [\texttt{sarcasm}]\\[4pt]
    \end{minipage}
    &
    \begin{minipage}[t]{0.94\linewidth}
      \raggedright
      \textbf{About:} The meme features a humorous and surreal scene with a penguin sitting on a chair in a formal setting, alongside a well-known public figure and another individual. The penguin is humorously demanding the removal of tariffs, threatening to defecate on the chair if its demands are not met. The public figure responds with a comment about the penguin's lack of clothing, while the other individual expresses concern for the furniture. This meme likely satirizes political negotiations and trade discussions, using the absurdity of a penguin in a formal meeting to highlight the sometimes ridiculous nature of such events.\\[4pt]
      \textbf{Origin:} The meme likely originated from a digitally altered image of a political meeting, with added captions for comedic effect. It draws on the tradition of using animals in unexpected contexts to create humor.\\[4pt]
      \textbf{Spread:} This type of meme typically spreads on social media platforms like Twitter, Reddit, and Instagram, where users share and remix content for comedic purposes.\\[4pt]
      \textbf{Culture:} The meme belongs to internet culture, specifically the subculture of political satire and absurdist humor.\\[4pt]
      \textbf{Structure:} The meme uses a photo with superimposed speech bubbles to create a dialogue between the characters.\\[4pt]
      \textbf{Usage:} It is used for satire and humor, often to mock political situations or negotiations.\\[4pt]
      \textbf{Target Audience:} Individuals interested in politics, satire, and internet humor.\\[4pt]
      \textbf{Emotional Impact:} The meme evokes humor and amusement through its absurdity and unexpected dialogue.\\[4pt]
      \textbf{Cognitive Load:} Understanding the meme requires some knowledge of political contexts and the use of animals in humor.\\[4pt]
      \textbf{Persuasive Technique:} The meme uses irony and absurdity to critique political negotiations.\\[4pt]
      \textbf{Safety Flag:} [\texttt{sarcasm}]
    \end{minipage}
  \end{tabularx}}
\end{minipage}}

\caption{
Qualitative comparison on the ``penguin–tariff in the Oval Office'' meme.
}
\label{fig:case_penguin_tariff}
\end{figure*}

\begin{figure*}[t]
\centering
\setlength{\fboxsep}{6pt}
\fcolorbox{black}{gray!5}{
\begin{minipage}{0.94\linewidth}
  \centering
  \includegraphics[width=0.15\linewidth]{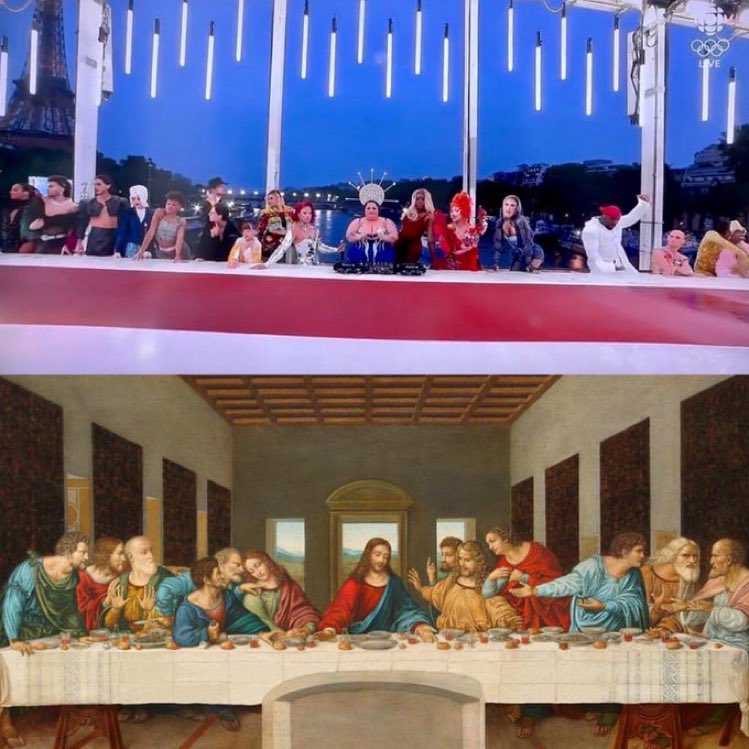}

  {\scriptsize
  \renewcommand{\arraystretch}{1.05}
  \begin{tabularx}{0.94\linewidth}{@{}X X X@{}}
    \textbf{(a) MemeTrace (Agent)} &
    \textbf{(b) Baseline (VLM)} &
    \textbf{(c) Robustness (bar\_top12 perturbation)} \\
    \begin{minipage}[t]{0.94\linewidth}
      \raggedright
      \textbf{About:} The meme juxtaposes a modern scene from the Paris Olympics Opening Ceremonies with Leonardo da Vinci’s \textit{The Last Supper}. The top image features a tableau with a central female figure and drag queens, resembling the iconic arrangement of figures in da Vinci’s painting. This scene was part of a performance on July 26, 2024, which drew criticism for its perceived mockery of Christianity. The French Bishops’ Conference and Bishop Robert Barron condemned it as a “gross mockery.” The performance was intended to promote inclusion and love, according to artistic director Thomas Jolly, but its resemblance to \textit{The Last Supper} sparked controversy. The scene included 18 drag performers and a man painted blue, evoking Dionysus, with the Eiffel Tower as a backdrop. The controversy highlights tensions between artistic expression and religious sensitivities.\\[4pt]
      \textbf{Origin:} The meme originated from the Paris Olympics Opening Ceremonies on July 26, 2024, where a performance resembling \textit{The Last Supper} took place.\\[4pt]
      \textbf{Spread:} The meme spread across social media platforms, fueled by the controversy and media coverage of the event.
    \end{minipage}
    &
    \begin{minipage}[t]{0.94\linewidth}
      \raggedright
      \textbf{About:} The meme juxtaposes a modern scene with a group of people, likely drag performers, mimicking the arrangement of figures in Leonardo da Vinci’s \textit{The Last Supper}. The top image shows individuals dressed in flamboyant costumes seated at a long table, reminiscent of the famous painting depicted in the bottom image. This visual parody is often used to humorously comment on contemporary culture by drawing parallels with iconic historical or religious imagery.\\[4pt]
      \textbf{Origin:} This type of meme likely originated from the widespread practice of parodying \textit{The Last Supper}, a well-known artwork by Leonardo da Vinci. Such parodies have been popular in meme culture for years, often used to satirize or comment on modern events or groups.\\[4pt]
      \textbf{Spread:} Memes like this typically spread across social media platforms such as Instagram, Twitter, and Facebook, where visual humor and cultural references are widely shared and appreciated.
    \end{minipage}
    &
    \begin{minipage}[t]{0.94\linewidth}
      \raggedright
      \textbf{About:} The meme juxtaposes a modern parody of Leonardo da Vinci’s \textit{The Last Supper} with the original painting. The top image features drag queens reenacting the famous scene, which was part of the 2024 Paris Olympic Games opening ceremony on July 28, 2024. This performance was intended as a parody and included a scene where Jesus was replaced with an obese woman, and queer and trans figures depicted apostles. The event drew criticism from figures like Mike Johnson, Speaker of the United States House of Representatives, and Marion Maréchal, a far-right politician, who viewed it as disrespectful to Christian beliefs. The Vatican also labeled it an “offense” to Christians. Despite the controversy, the artists involved claimed no intent to offend, highlighting cultural tensions between artistic expression and religious sensitivities. The meme reflects broader societal debates on freedom of expression and respect for religious symbols.\\[4pt]
      \textbf{Origin:} The meme originated from the 2024 Paris Olympic Games opening ceremony on July 28, 2024, where drag queens performed a parody of \textit{The Last Supper}.\\[4pt]
      \textbf{Spread:} The meme spread across social media platforms, particularly on X, where it became part of the broader meme culture surrounding the Paris 2024 Olympics.
    \end{minipage}
  \end{tabularx}}
\end{minipage}}

\caption{
Qualitative comparison under different system configurations for the ``Last Supper'' meme.
}
\label{fig:case_lastsupper_paris2024}
\end{figure*}

We present three representative examples comparing MemeTrace and the VLM baseline on the meme provenance task (Figures~\ref{fig:case_img33_immigration},~\ref{fig:case_penguin_tariff},  and~\ref{fig:case_lastsupper_paris2024}), where the final case additionally shows its report under UI overlay perturbation. 
Each case includes the meme images at the top, followed by structured reports generated by MemeTrace and the baseline.

\mypara{(a) ``Immigration/Florida–turnpike'' Meme}  
This meme refers to a 2025 traffic accident in Florida, in which an undocumented Indian immigrant caused the deaths of three people. It satirizes the immigration debate and political polarization in the United States. MemeTrace correctly identifies the incident and recognizes the background of the driver, and explains the controversies surrounding the accident. In contrast, the VLM baseline merely repeats the visible text and imagery, failing to capture the underlying context.

\mypara{(b) ``Penguin–Tariff'' Meme}  
This political satire shows Donald Trump negotiating tariffs with a penguin in the Oval Office. MemeTrace successfully locates the April 2025 tariff-policy event as its contextual background, explains the meme’s use of absurd humor to critique U.S. trade decisions, and identifies the related ``penguin-tariff'' trend on X/Twitter. In contrast, the baseline merely paraphrases the dialogue in the meme, and fails to recognize its geopolitical context, highlighting MemeTrace’s strength in factual grounding and multimodal understanding of political satire.

\mypara{(c) ``Last Supper Parody'' Meme}  
MemeTrace successfully identifies that this meme is associated with the 2024 Paris Olympics opening-ceremony tableau and Leonardo da Vinci’s The Last Supper, and further points out the religious controversies and public reactions triggered by the ceremony. The baseline only describes the visual similarity between the two scenes, without recognizing the context of the 2024 Paris Olympics opening ceremony. Even under moderate UI overlay interference, MemeTrace retains key interpretive details, demonstrating its robustness.

Across all three examples, MemeTrace provides analyses that are richer in factual detail and more contextually grounded than those of the baseline, confirming its effectiveness in producing coherent provenance reports.

\section{Safety-Flag Analysis}
\label{appendix:safety}
\begin{figure}[t]

  \centering
 
  \subfloat[Overall distribution of safety flags.]
  {\includegraphics[width=0.85\linewidth]{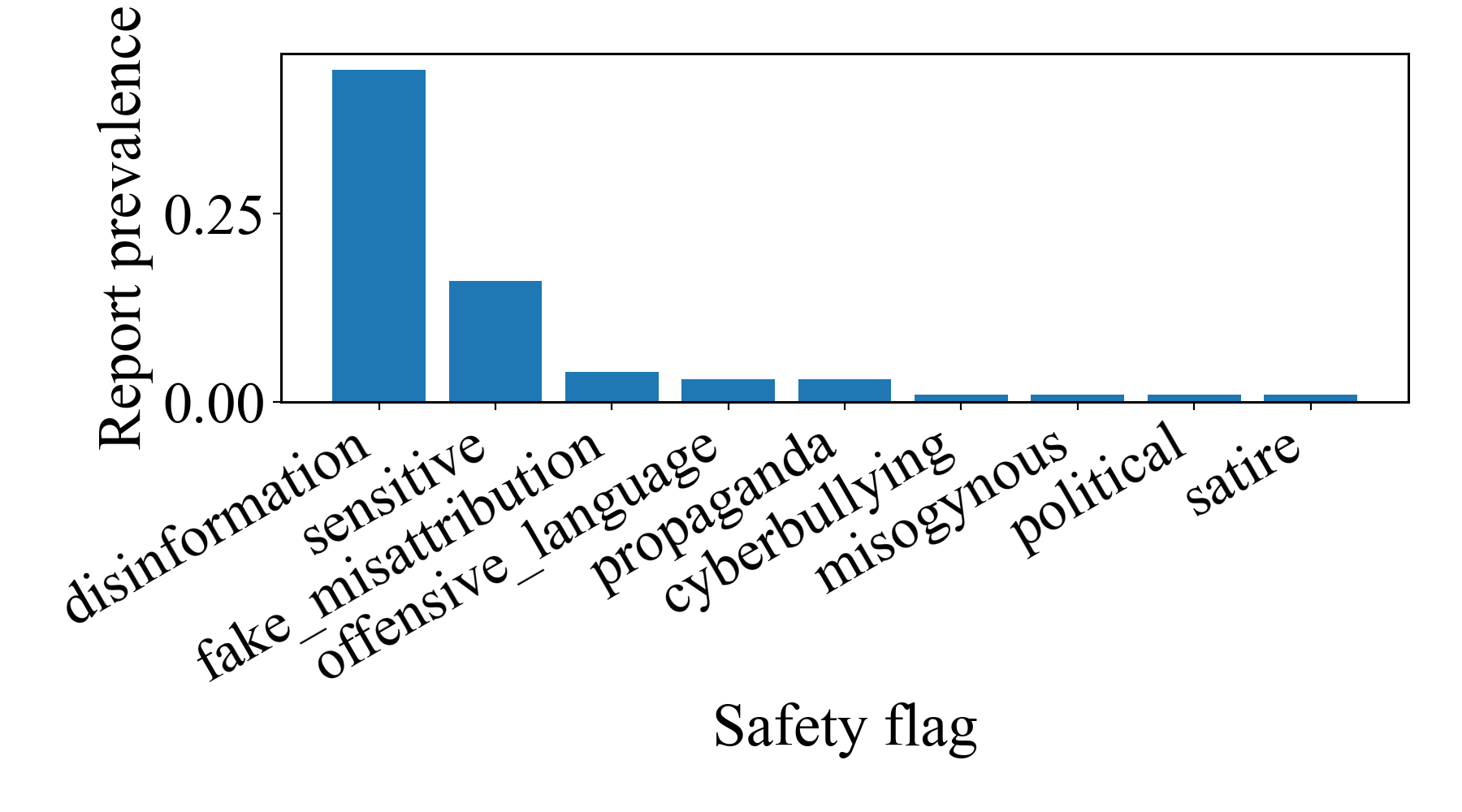}
   \label{fig:safety_flag_prevalence}}

  \subfloat[Topic-wise distribution of \textit{disinformation} and \textit{sensitive} flags.]
  {\includegraphics[width=0.85\linewidth]{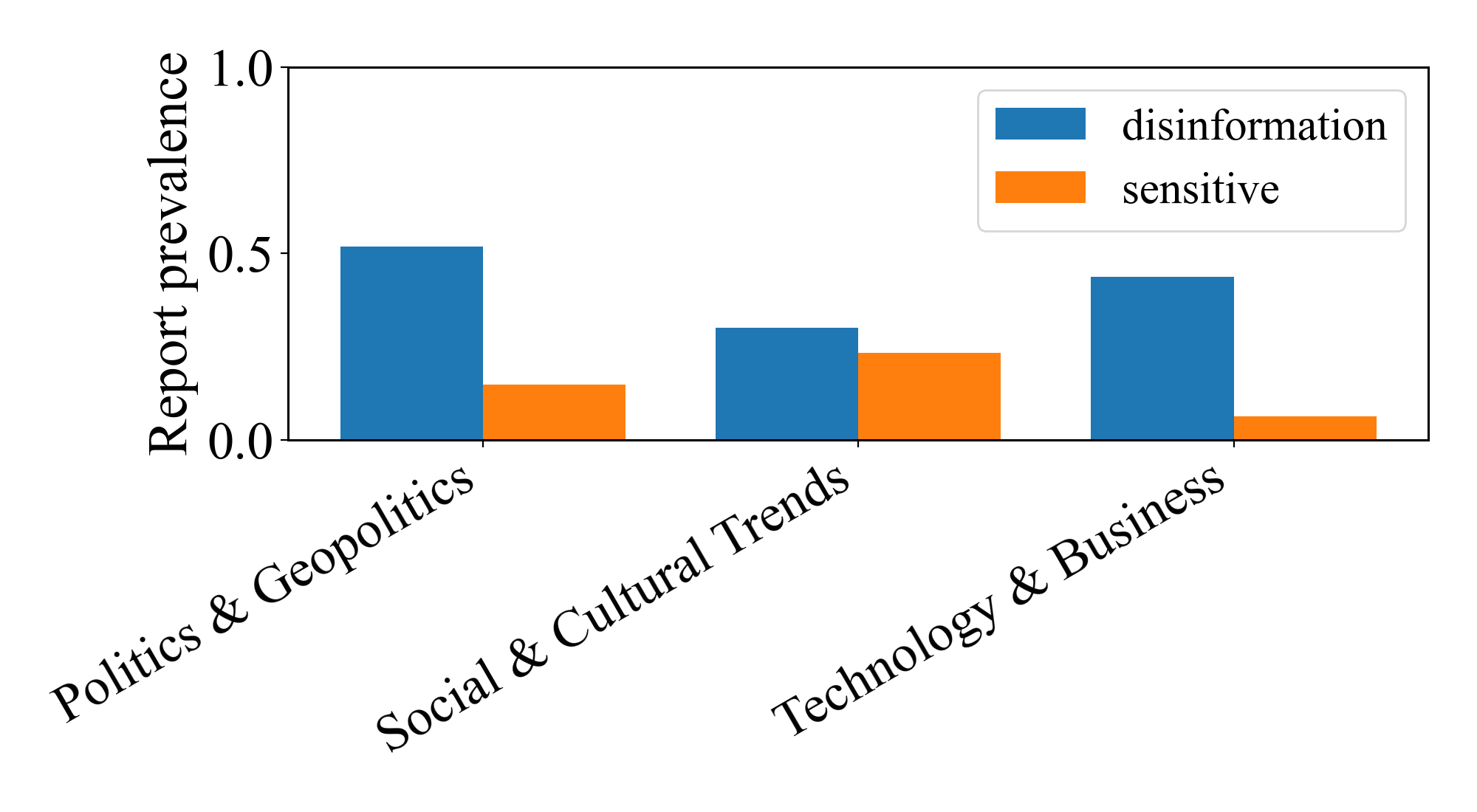}
   \label{fig:topic_disinfo_sensitive_prevalence}}

  \subfloat[Co-occurrence matrix of safety flags.]
  {\includegraphics[width=0.8\linewidth]{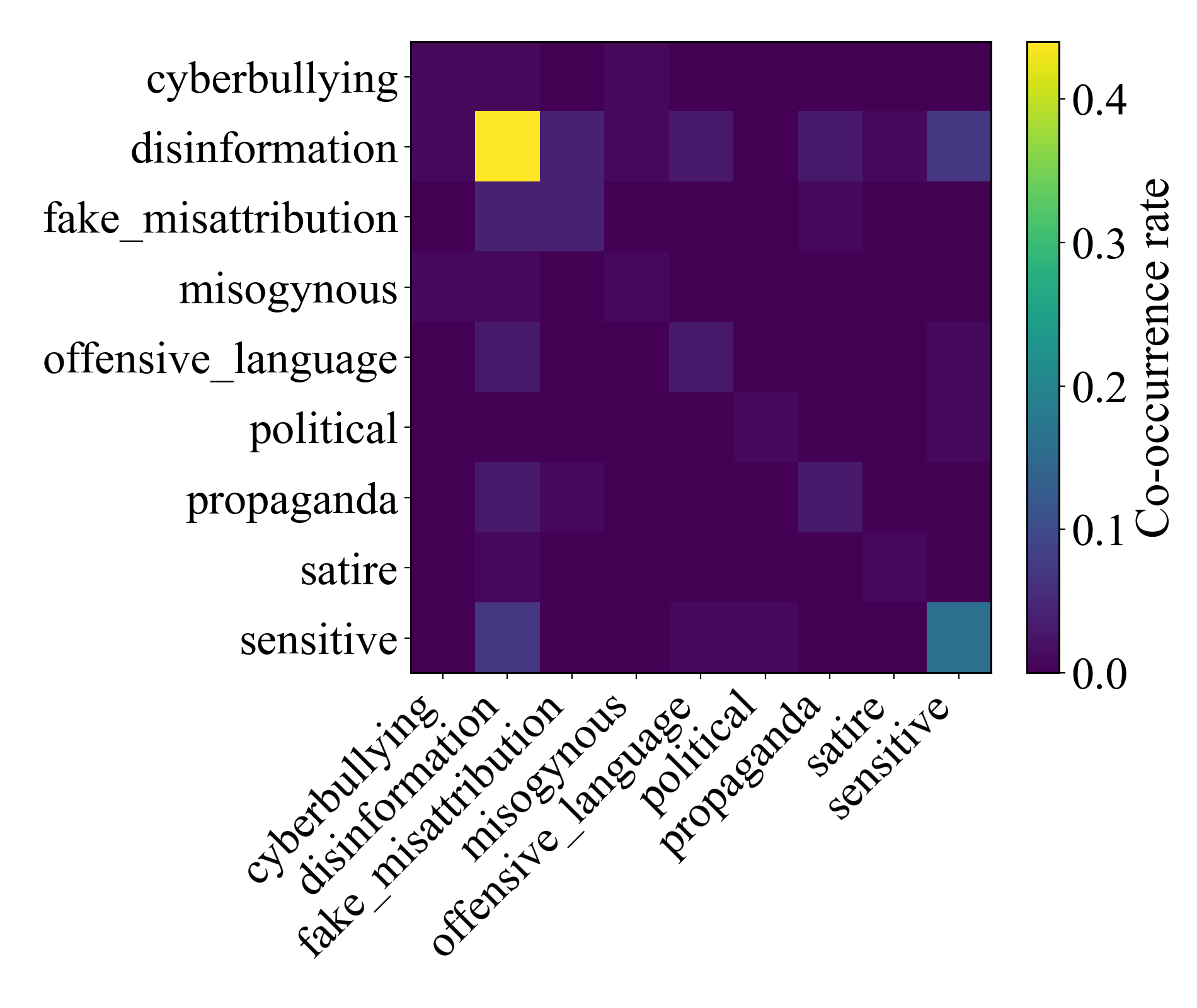}
   \label{fig:safety_flag_cooccurrence_heatmap}}

  \caption{
      Safety-flag statistics.
  }
  \label{fig:safety_flag_analysis}
\end{figure}

\autoref{fig:safety_flag_analysis} presents extended safety-flag statistics, including their overall distributions, topic-wise distributions, and co-occurrence matrix.

\section{Trace-Log Visualizations}
\label{appendix:trace_log}
\begin{figure}[t]
    \centering
    \setlength{\fboxsep}{6pt}
    \fcolorbox{black}{gray!5}{
        \includegraphics[width=0.75\linewidth]{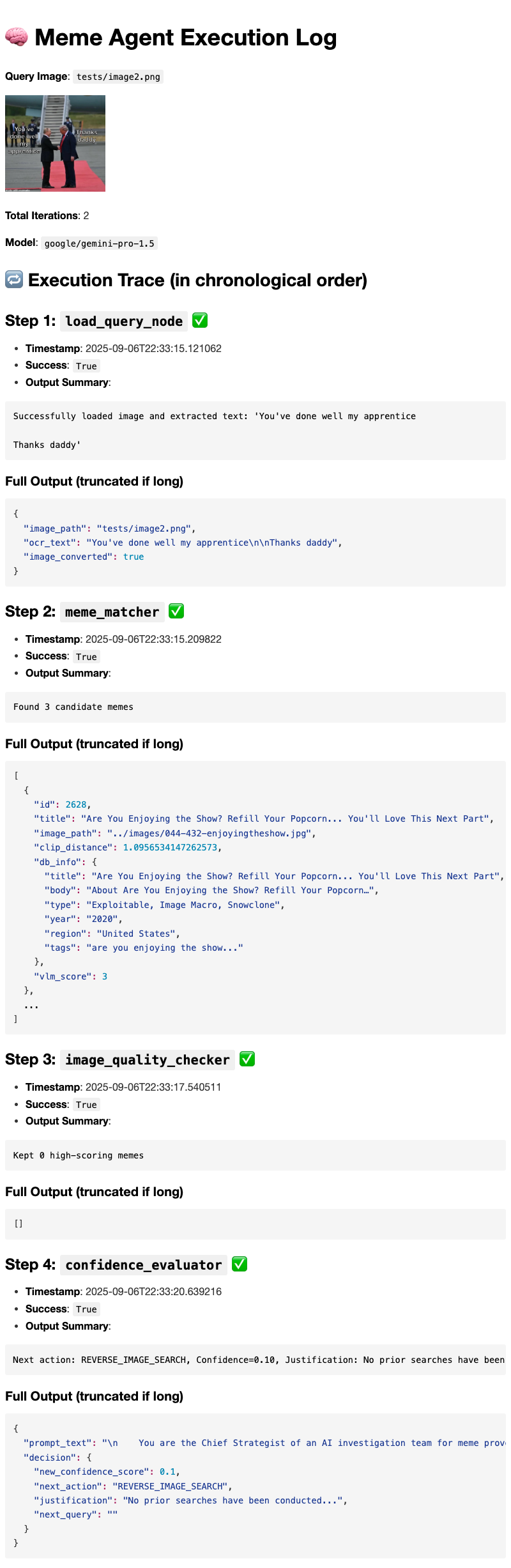}
    }
    \caption{
      Example execution log trace from MemeTrace (Gemini~1.5~Pro) on \textit{memeid = 2}.
      Some intermediate steps have been omitted for brevity.
    }
    \label{fig:trace_log_example}
\end{figure}

To illustrate the system's execution process, we record runtime logs and save them in Markdown format for convenient visualization.
\autoref{fig:trace_log_example} shows an example trace, demonstrating the first several steps of the system's execution flow.
All steps are logged in detail, ensuring that the reasoning process is transparent and auditable.

\begin{figure}[t]
    \centering
    \includegraphics[width=\linewidth]{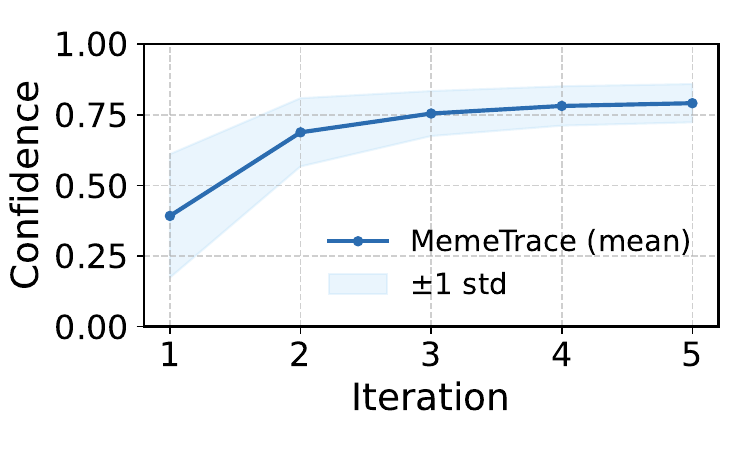}
    \caption{
    Confidence evolution averaged over the main MemeTrace experiments ($n{=}100$).
    The solid line represents the mean confidence, and the shaded region denotes $\pm$1 standard deviation.
    }
    \label{fig:trace_confidence_curve}
\end{figure}

We plot the average confidence progression during iterative retrieval, as shown in \autoref{fig:trace_confidence_curve}.
The results indicate that the mean confidence increases steadily with each iteration and stabilizes around the fifth step, supporting our choice of setting the maximum iteration limit to five.

\section{System Efficiency and Cost Analysis}
\label{appendix:system_efficiency}
\subsection{Execution Profiles}
Based on execution logs, MemeTrace performs an average of 18.46 processing steps per meme. These include 11.54 multimodal LLM API calls, 3.92 external search API calls (2.92 via Tavily and 1.00 via SerpApi), and approximately 3 local operations such as OCR processing and database retrieval.
We observe that the majority of latency and API cost arises from LLM-based iterative reasoning.

\subsection{Cost and Latency Analysis}
In our experiments, MemeTrace incurs an average cost of \$0.1259 per meme, with an end-to-end processing time of approximately 111 seconds per query.
While this latency is higher than that of lightweight baselines, it remains substantially lower than manual meme provenance analysis, which typically takes tens of minutes of expert effort per meme.

\subsection{Practical Usage Scenarios}
In realistic moderation pipelines, MemeTrace can be integrated as a selective fallback module, invoked when lightweight models or rule-based systems fail to reach confident decisions for challenging or ambiguous cases.

\end{document}